\documentclass[10pt,journal,compsoc]{IEEEtran}
\usepackage{cite}
\usepackage{amsmath,amssymb,amsfonts}
\usepackage{graphicx}
\usepackage{textcomp}
\def\BibTeX{{\rm B\kern-.05em{\sc i\kern-.025em b}\kern-.08em
    T\kern-.1667em\lower.7ex\hbox{E}\kern-.125emX}}

\usepackage{subfigure}
\usepackage[ruled, linesnumbered]{algorithm2e}
\usepackage{wrapfig}
\usepackage{multirow}
\usepackage{url}
\usepackage{booktabs}
\usepackage{threeparttable}
\usepackage[colorlinks=true, allcolors=blue]{hyperref}

\newcommand\blue[1]{\color{black}#1}

\PassOptionsToPackage{dvipsnames}{xcolor}
\usepackage{color}
\usepackage{tikz}
\usetikzlibrary{arrows,positioning,fit,petri,shapes,backgrounds,decorations.pathmorphing,calc,shapes.misc, arrows, decorations.markings}

\newcommand{\eg}{e.\,g.,\ }
\newcommand{\ie}{i.\,e.,\ }

\newcommand{\et}{{et al.}}
\newcommand{\mb}{\mathbf}
\newcommand{\mc}{\mathcal}

\ifCLASSINFOpdf
\else
\fi
\begin{document}
%
\title{EmoBed: Crossmodal Emotion Embedding \\for Emotion Recognition}
\title{Enhancing Monomodal Emotion Recognition via Audiovisual Signals}
\title{Strengthening Monomodal Emotion Recognition with Implicit Fusion of Audiovisual Signals}
\title{EmoBed: Strengthening Monomodal Emotion Recognition by Exploring Audiovisual Signals}
\title{EmoBed: Strengthening {\blue Monomodal} Emotion Recognition via Training with Crossmodal Emotion Embeddings}

\author{Jing~Han,
        Zixing~Zhang,
        Zhao~Ren, 
        and~Bj\"orn~Schuller
\IEEEcompsocitemizethanks{\IEEEcompsocthanksitem J.~Han, Z.~Ren, and B.~Schuller are with the ZD.B Chair of Embedded Intelligence for Health Care and Wellbeing, University of Augsburg, 86159 Augsburg, Germany. E-mail: \{jing.han, zhao.ren\}@informatik.uni-augsburg.de
\IEEEcompsocthanksitem Z.~Zhang and B.~Schuller are with Group on Language, Audio \& Music, Imperial College London, London SW7 2AZ, UK. E-mail: \{zixing.zhang, bjoern.schuller\}@imperial.ac.uk}
}

\IEEEtitleabstractindextext{%
\begin{abstract}
Despite remarkable advances in emotion recognition, they are severely restrained from either the essentially limited property of the employed single modality, or the synchronous presence of all involved multiple modalities. Motivated by this, we propose a novel crossmodal emotion embedding framework called EmoBed, which aims to leverage the knowledge from other auxiliary modalities to improve the performance of an emotion recognition system at hand. The framework generally includes two main learning components, \ie joint multimodal training and crossmodal training. Both of them tend to explore the underlying semantic emotion information but with a shared recognition network or with a shared emotion embedding space, respectively. In doing this, the enhanced system trained with this approach can efficiently make use of the complementary information from other modalities. Nevertheless, the presence of these auxiliary modalities is not demanded during inference. To empirically investigate the effectiveness and robustness of the proposed framework, we perform extensive experiments on the two benchmark databases RECOLA and OMG-Emotion for the tasks of dimensional emotion regression and categorical emotion classification, respectively. The obtained results show that the proposed framework significantly outperforms related  baselines {\blue in monomodal inference}, and are also competitive or superior to the recently reported systems, which emphasises the importance of the proposed crossmodal learning for emotion recognition. 
\end{abstract}

}

\maketitle

\IEEEdisplaynontitleabstractindextext

%
\IEEEpeerreviewmaketitle

\IEEEdisplaynontitleabstractindextext

%
\IEEEpeerreviewmaketitle

\IEEEraisesectionheading{\section{Introduction}\label{sec:introduction}}

\IEEEPARstart{A}{utomatic} emotion recognition endows machines with the capability of natural and empathic communication with humans. Thus, it is considered to be essential to sustain long-term human--machine interactions and is even critical in shifting the present artificial intelligence into the next generation enhanced with emotional characteristics~\cite{Cowie01-Emotion,Zeng09-Survey,Zhang14-Distributing,Cambria16-Affective}. Over the past decades, significant advances have been made to improve the accuracy and robustness of emotion recognition systems in either a monomodal or multimodal scenario. 

The {\em monomodal} emotion recognition systems normally independently explore the prominent features for the emotions of interest, from one specific modality, such as audio, video, image, text, or physiology~\cite{El11-Survey, Happy15-Automatic, Clavel16-Sentiment, Alarcao18-Emotions}. With the advent of deep learning, such modality-specific systems have continually achieved promising performance~\cite{Zhang18-Cross,Han18-Adversarial,Mohammed18-DAF}. In contrast, the {\em multimodal} systems tend to jointly utilise several modalities, with an aim to take advantage of complementary or supplementary information from different media cues~\cite{Zeng09-Survey,Nicolaou11-Continuous,Poria16-Fusing}. Benefited from this, such systems have been consistently evaluated to be superior to the monomodal systems in numerous previous works~\cite{Han17-Strength,Zhang18-Dynamic,Alghowinem18-Multimodal}.  

Albeit the notable advantages, in the evaluation phase, most of the multimodal emotion recognition systems require the synchronous presence of all modalities that are employed in the previous training phase~\cite{Poria16-Fusing,Nicolaou11-Continuous, Han17-Strength,Zhang18-Dynamic,Alghowinem18-Multimodal}. This severely impedes their application in real life, since  it is a common case that information from some particular modalities is missing. For example, a camera could be not always fixed in front of a user, or could not work in darkness, which results in invalid or missing visual signals. Likewise, a user could be silent although she/he is emotional, leading to the missing of audio data. The absence of any involved modalities often leads to corruption or performance degradation of a pre-trained multimodal system~\cite{Zeng09-Survey}. 

A straightforward way to address this issue often makes use of the integration of an additional component, such as a voice activity detector and a face detector, in front of the multimodal recognition systems~\cite{Zeng09-Survey}. Once the absence of a particular modality is detected, the prediction process can be automatically re-directed to another system that is trained via an accordingly reduced number of modalities. Nevertheless, such a system is normally inferior to the system with all modalities as aforementioned. 


To embrace the advantages and avoid the shortages of both systems, in this article, we propose a novel {\em crossmodal Emotion emBedding} framework, namely {\em EmoBed}. 
{\blue The underlying idea of this framework is to transfer the knowledge from other auxiliary modalities to a target modality, in order to enhance the performance of a monomodal emotion recognition model.  Basically, it consists of two main processes: the {\em joint multimodal training} process and the {\em crossmodal training} process.  The former process utilises the data from multiple modalities to jointly train a shared network, with an assumption that the knowledge from different modalities could be implicitly transferred to or fused by the network. Meanwhile,  the {\blue latter} process utilises a triplet loss~\cite{Chechik10-Large,Schroff15-Facenet} to minimise the distance of intra-class representations while maximising the inter-class ones, regardless of their corresponding modality types. In doing this, it forces the extracted high-level representations cross modalities shared a similar space, where the intra-class representations have a close distance while the inter-class ones have a long distance. 
}

{\blue This framework holds two main advantages compared with traditional multimodal emotion recognition systems~\cite{Poria16-Fusing,Nicolaou11-Continuous, Han17-Strength,Zhang18-Dynamic,Alghowinem18-Multimodal}. Firstly, it only requires the data from auxiliary modalities in the training stage, where the knowledge is supposed to be transferred to the target modality. In the inference stage, the auxiliary modalities are not needed anymore. Therefore, it overcomes the synchronous presence problem of traditional multimodal systems. Secondly, when training the network, it does not demand a paired data. That is, the signals from different modalities are unnecessarily time-aligned. Thus, the data from heterogeneous corpora (\ie modality mismatch) can be used for training our framework. This advantage largely releases the signal-alignment requirement of traditional multimodal systems.}


Furthermore, our work is partially inspired by the multi-task learning paradigm, where multiple tasks are jointly trained with a shared network and several task-specific networks. It has been repeatedly demonstrated including in affective computing that such a learning process can lead to a better generalisation of the representations learnt from the shared networks~\cite{Eyben12-MAC, Xia17-AMT, Han17-From, Zhao18-MMM, Taylor18-Personalized}. 
Similarly, in this work, we assume that multiple modalities could also benefit for training a monomodal framework, through optimising the parameters of a shared network.

Overall, the major contributions of this work include: (i) We propose a novel learning framework -- EmoBed -- to explore knowledge from auxiliary modalities for an emotion recognition system; (ii) we jointly train the network with the heterogeneous data, which, however, is unnecessary to be paired; (iii) we extract modality-invariant emotion embeddings in a latent space via a triplet loss. Although triplet loss has been implemented in the emotion recognition literature~\cite{Han18-Emotion,Huang18-Speech}, it was merely utilised to distil discriminative representations within speech signals, which differs from the proposed work that aims to distil the modality-invariant emotion embeddings; finally, (iv) we comprehensively investigate the effectiveness and robustness of the model for both dimensional continuous emotion regression and categorical discrete emotion classification. 
Note that, for the sake of clarification, we define the traditional emotion recognition system without any crossmodal training as a {\em classic} emotion recognition system, whereas the system enhanced by our proposed crossmodal technologies as an {\em enhanced} emotion recognition system. 
Despite the fact that our proposed approaches can be used for more than one modality, for the sake of simplicity, in the present article, we mainly focus on the visual and audio modalities for emotion recognition, as cameras and microphones are pervasive in the world. Thus, the audio-only or video-only based system is assumed to be monomodal system. 

The remainder of this article is structured as follows. 
First, we brief related works for crossmodal and multimodal emotion recognition systems in Sections~\ref{sec:relatedwork}. 
After that, in Section~\ref{sec:method}, we elaborately describe the proposed EmoBed framework in detail. Then, in Section~\ref{sec:ex} and~\ref{sec:res}, we perform comprehensive experiments on two audiovisual emotional databases, to assess the performance of the proposed approach in both emotion regression and classification tasks. Finally, in Section~\ref{sec:conclusion}, we draw conclusions and point out some promising related research directions.

\section{Related Work}
\label{sec:relatedwork}
In this section, we summarise the most related and recently reported works on the crossmodal training and multimodal emotion recognition systems for emotion recognition, respectively.

\begin{figure*}[!ht]
    \centering
    \includegraphics[height=2.3in,trim={0.0cm 0.0cm 0.0cm 0.0cm},clip]{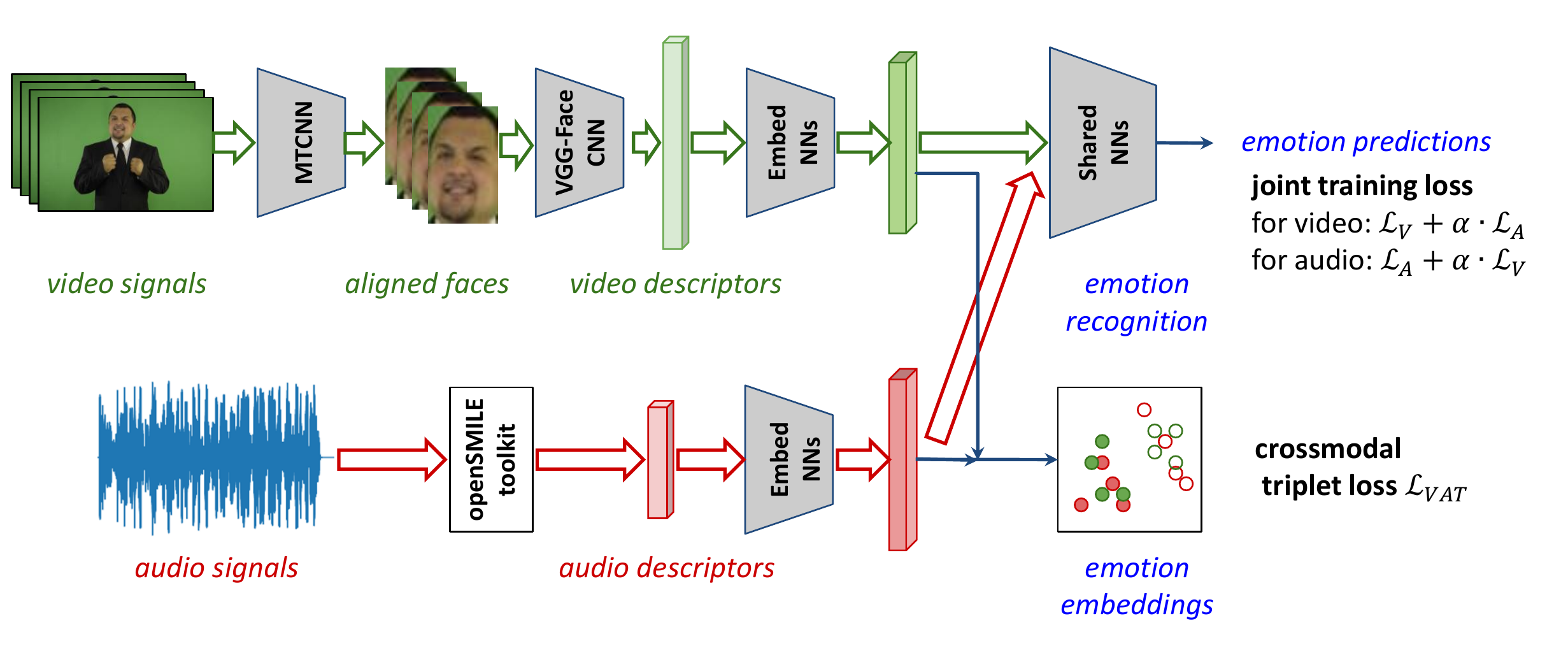}
    \caption{{\blue The proposed crossmodal Emotion emBedding (EmoBed) framework for monomodal emotion recognition.}}
    \label{fig:training}
\end{figure*}

\subsection{Crossmodal Training}
\label{subsec:rw_cross}
Recently, there have been an increasing number of studies to investigate approaches to transfer knowledge across domains or modalities~\cite{Yang15-CFL, Kang15-LCF, Aytar16-SLS, Wang17-ACM}. One notable relevant work is SoundNet~\cite{Aytar16-SLS} for large-scale natural sound representation learning under a student-teacher training procedure. In particular, a teacher vision network can guide the student sound network to recognise acoustic events in an unsupervised learning manner. During training, KL-divergence between the posterior probabilities of the teacher and student networks is minimised, and thus, knowledge is transferred from well established visual recognition models into audio ones. Of more relevance is a recent work {\blue by Albanie~\et~\cite{Albanie18-ERS}}, 
where the authors proposed to produce aligned embeddings for speech emotion recognition, by distilling the knowledge of a strong teacher network for facial emotion recognition.

Our approach differs from {\blue the one proposed by Albanie~\et~\cite{Albanie18-ERS}} 
mainly in two key aspects. {\blue Firstly}, although the student network is learnt without access to any form of labelled audio, the main limitation of the approach is that it requires a well-trained complex teacher network, which arguably has not yet been developed in emotion recognition. In our method, however, the network is trained from scratch. {\blue Second}, the results reported {\blue in the work~\cite{Albanie18-ERS}} indicate that the performance of the student network falls short of the performance of the teacher network, and cannot compete with a fully supervised network. In contrast, our primary focus goes to provide additional knowledge from a second modality to assist the targeted modality. As a consequence, the performance of our approach is expected to be superior to a fully supervised monomodal network. 

\subsection{Multimodal Emotion Recognition}
\label{subsec:rw_multi}
As mentioned in Section~\ref{sec:introduction}, multimodal fusion approaches have been widely conducted to exploit {\blue complementary information}, to improve the performance and robustness of emotion recognition systems measurably~\cite{Han17-Strength, Metallinou12-Context}. To this aim, various data fusion strategies have been extensively utilised, and they normally can be classified into three categories, namely, feature-level fusion, decision-level, and model-level fusion~\cite{Zeng09-Survey}. Typically, feature-level fusion ({also known as} early fusion) straightforwardly concatenates audio and visual features into one combined feature vector, which is then used as the input for modelling~\cite{Busso04-AOE, Wimmer08-LFO}. 
In contrast, decision-level fusion ({also known as} late fusion) combines the predictions, rather than the features, from the modality-specific models for a final decision by the use of certain suitable criteria~\cite{Busso04-AOE, Metallinou10-DLC}. 
{\blue In addition,} model-level fusion fuses the intermediate representations instead, and in this manner learns to model the potential hidden correlations among features~\cite{Zeng05-AAR,Han17-Strength,Han17-Prediction}. 
{\blue Compared with the former two fusion strategies, model-level fusion is supposedly more bio-inspired from the cognitive perspective~\cite{Toprak18-Evaluating}. Overall,}
these three fusion strategies are practical and helpful, to different extents, for audiovisual emotion recognition.

However, these approaches focus on multimodal scenarios and rely heavily on the existence of signals from all sensors. 
In contrast to these works, we exploit the hidden correlation of multiple modalities in an implicit fusion manner, and thus it later can be implemented in a more flexible setting, as information from auxiliary modalities is not required during inference.
\section{Crossmodal Emotion Embedding System}
\label{sec:method}

In this work, we aim to attain a shared embedding space to explore the latent correlation between audio and video signals for monomodal emotion recognition, and the training stage of the proposed EmoBed framework is depicted in Fig.~\ref{fig:training}.
Typically, after extracting audio and video descriptors via several standard and essential processing steps, we jointly train two {\blue modality-specific} 
networks to project these multimodal descriptors into a common space, the representations of which can then be applied to predict emotions. 

Mathematically, the two embedding functions $f_A:\mathbb{R}^M \rightarrow \mathbb{R}^E$ and $f_V:\mathbb{R}^N \rightarrow \mathbb{R}^E$, aim at mapping audio inputs in $\mathbb{R}^M$ and visual inputs in $\mathbb{R}^N$ onto embedded representations accordingly in a shared coordinate space $\mathbb{R}^E$. To learn such embedding functions, we first introduce joint training with audiovisual data in Section~\ref{subsec:method_joint}. Moreover, to learn useful semantic representations, we further employ crossmodal triplet loss in the learning process in Section~\ref{subsec:method_mono}. Lastly, the proposed EmoBed framework is given in Section~\ref{subsec:method_cross}, by integrating the merits of the joint training and the crossmodal training.

\subsection{Joint Training with Audiovisual Data}
\label{subsec:method_joint}

\begin{figure*}[t]
    \centering
    \includegraphics[height=2.0in,trim={0.2cm 5.05cm 0cm 1.6cm},clip]{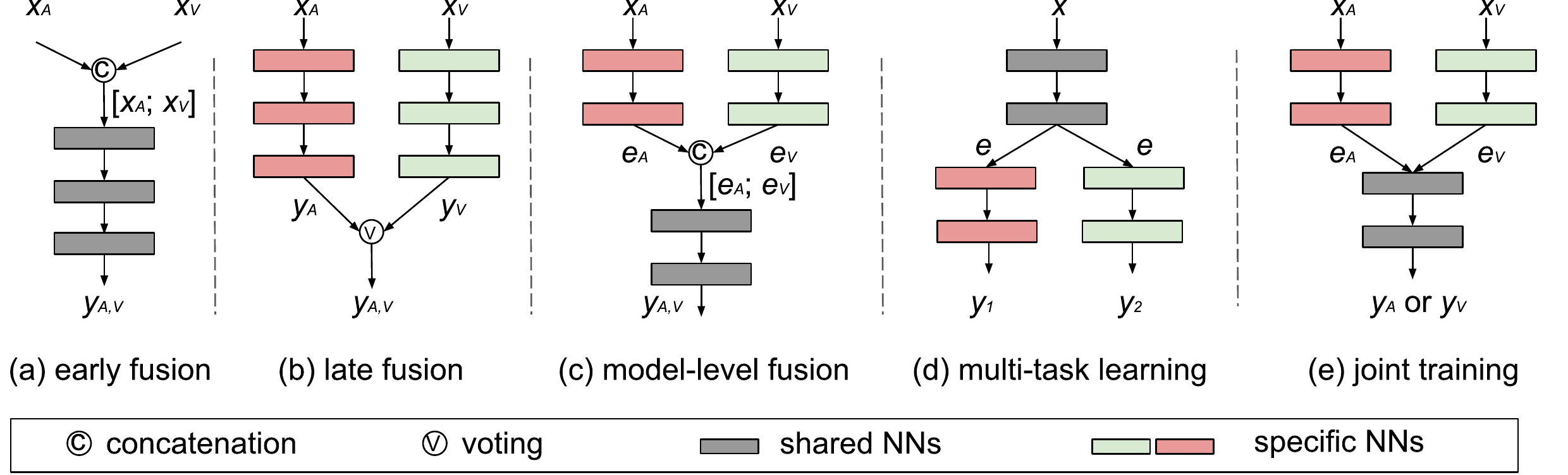}
    \caption{Structure comparison among the proposed joint audiovisual training (e), and other related multimodal learning frameworks (\ie early fusion (a), late fusion (b), model-level fusion (c)), and multi-task learning (d). }
    \label{fig:overview}
\end{figure*}

In this subsection, we demonstrate how to learn a common embedding space for monomodal emotion recognition with joint training loss using audiovisual emotional data.
Before that, we first briefly differentiate our joint training from other related structures in the following paragraphs, as depicted in Fig.~\ref{fig:overview}.

Three conventional multimodal fusion paradigms are demonstrated in Fig.~\ref{fig:overview}(a)-(c). Note that, although information of multiple modalities is fused at different levels, they all contribute to multimodal emotion recognition systems.
Concretely, given $\mb{x}_{(\cdot)}$, $\mb{e}_{(\cdot)}$, and $y_{(\cdot)}$ denoting the monomodal input feature, the learnt hidden-layer representation, and the output prediction for audio $A$ and video $V$, respectively, the combination of audio and video knowledge is in forms of $[\mb{x}_A; \mb{x}_V]$ for feature-level fusion, weighted averaging based on $y_A$ and $y_V$ for decision-level fusion, or $[\mb{e}_A; \mb{e}_V]$ for model-level fusion. It should be noted that, these models can be utilised, if and only if both $\mb{x}_A$ and $\mb{x}_V$ are available as inputs of the model, and there is no need for $\mb{e}_A$ and $\mb{e}_V$ to be of the same dimensions. In the proposed joint training model, albeit the constraint of the existing of both $\mb{x}_A$ and $\mb{x}_V$ remains during training, the model can then be applied under a monomodal setting.

{\blue Furthermore,} our model is {\blue also dissimilar to} multi-task learning, which is illustrated in Fig.~\ref{fig:overview}(d).
In multi-task learning, during the training phase,  an auxiliary task benefits the main task by updating the parameters in the shared front-end feature-learning network. In contrast, in the proposed model, we assume inputs of an auxiliary modality can help improve the emotion prediction of the main modality, by optimising the parameters of the shared back-end predicting network.

Let us denote an audio feature vector as $\mb{x}_A\in \mathbb{R}^M$ and its corresponding visual feature vector as $\mb{x}_V\in \mathbb{R}^N$, where $M$ and $N$ are the dimensions of the audio and visual vectors, respectively. As depicted in Fig.~\ref{fig:overview}(e), $\mb{x}_A$ and $\mb{x}_V$ are fed into two modality-specific subnetworks, the process of which can be formulated as follows:
 \begin{equation}
  \label{eq-map}
  \mb{e}_A = f_A(\mb{x}_A),\, \mb{e}_V = f_V(\mb{x}_V),
  \end{equation}
where the function $f_A(\cdot):\mathbb{R}^M \rightarrow \mathbb{R}^E$ and the function $f_V(\cdot):\mathbb{R}^N \rightarrow \mathbb{R}^E$ map each input of different modalities into the same subspace, resulting in corresponding $E$-dimensional representations $\mb{e}_A$ and $\mb{e}_V$.
After that, the following shared layers are applied to estimate the final predictions, and this process can be formulated as follows:
 \begin{equation}
  \label{eq-predict}
  y_A = f(\mb{e}_A),\, y_V = f(\mb{e}_V),
  \end{equation}
where the function $f(\cdot):\mathbb{R}^E \rightarrow \mathbb{R}$ estimates final predictions $y_A$ and $y_V$, separately.  

To efficiently aggregate the advantages of different modalities for monomodal emotion recognition (\ie speech emotion recognition or facial emotion recognition), the model is trained with a set of audiovisual features $\{(\mb{x}_A, \mb{x}_V)\}$. When the model is applied for speech emotion recognition, the joint loss function $\mc{J}(\theta)$ is calculated by:
  \begin{equation}
  \label{eq:fuse1}
   \mc{J}(\theta)=\mc{L}_A+\alpha \cdot \mc{L}_V,
  \end{equation}
where $\theta$ denotes the network parameters to be optimised, $\mc{L}_A$ and $\mc{L}_V$ stand for the loss of audio and video data, respectively, and $\alpha$ denotes the weight of the video prediction loss to regulate its contribution to $\mc{J}(\theta)$. The term $\alpha \cdot \mc{L}_V$ enforces the optimisation to take the auxiliary modality information into account. Similarly, for facial emotion recognition, the joint loss function in Eq.~(\ref{eq:fuse1}) is altered into
  \begin{equation}
  \label{eq:fuse2}
   \mc{J}(\theta)=\mc{L}_V+\alpha \cdot \mc{L}_A.
  \end{equation}
Moreover, the value of $\alpha$ is optimised on the development set, by achieving the best performance for the selected modality.

\subsection{Crossmodal Triplet}
\label{subsec:method_mono}
The primary focus of this subsection is to learn emotion-discriminative embeddings using crossmodal data via {\blue a} triplet loss function. 
In general, triplet loss forces to project the original descriptors into a latent space where instances with similar semantics are pulled together while instances with dissimilar semantics are pushed away. Consequently, the similarity of instances with the same semantic information is preserved in the learnt representations.
While the triplet constraint has been investigated intensively for images and text~\cite{hoffer15-DML, huang2013learning}, they are now starting to gain traction in audio and video studies~\cite{Han18-Emotion, Ding18-TEC}. Motivated by these successes, we propose a crossmodal triplet framework by adopting crossmodal triplet loss to supervise the learning process.

To form a triplet $tri=\{\mb{e},\mb{e}^+,\mb{e}^-\}$, for the embedding of a given instance $\mb{e}$, we select embeddings of another two instances, \ie $\mb{e}^+$ and $\mb{e}^-$.
In particular, $\{\mb{e}, \mb{e}^+\}$ (denoted as a \textit{positive pair}) are embeddings from the same class, with $\{\mb{e}, \mb{e}^-\}$ (denoted as a \textit{negative pair}) belonging to different classes.
To this end, we first calculate the semantic similarity of paired instances over a batch of embeddings.
Typically, given a pair of embeddings $\{\mb{e}, \mb{e}'\}$, their similarity $D$ can be computed as follows:
\begin{align}
D_{\mb{e},\mb{e}'} = \| \mb{e} - \mb{e}' \|_2,
\end{align}
where $\left \| \cdot \right \|_2$ denotes the Euclidean distance between two embeddings in the pair. Therefore, a pairwise Euclidean distance matrix can be generated by computing the distance between all embeddings. Note that, the diagonal of the obtained matrix is null, as the distance between one embedding and itself is zero. 

Subsequently, for each $\mb{e}$, we explore the distance matrix to find $\mb{e}^+$ and $\mb{e}^-$ from the batch, to form the hardest positive pair $\{\mb{e}, \mb{e}^+\}$ and the hardest negative pair $\{\mb{e}, \mb{e}^-\}$, by considering their emotion labels accordingly.
Particularly, $\mb{e}^+$ is one embedding which has the maximum distance from $\mb{e}$, among all embeddings having the same label as $\mb{e}$.
Conversely, $\mb{e}^-$ is another embedding which has the minimum distance from $\mb{e}$, among all embeddings with different labels from $\mb{e}$.

Once $\mb{e}^+$ and $\mb{e}^-$ are obtained and applied together with $\mb{e}$ to construct the hard triplet for each embedding, the triplet loss  constraint $\mc{L}_{T}$ can be estimated over all hard triplets by
\begin{equation}
    \mc{L}_{T} = \sum (D_{\mb{e}_i, \mb{e}_i^+} - D_{\mb{e}_i, \mb{e}_i^-}).
\end{equation}

Note that, to compute the crossmodal triplet loss $\mc{L}_{VAT}$, we first {\blue combine} the audio embeddings $\mb{e}_A$ and the video embeddings $\mb{e}_V$, to form a double-sized batch of embeddings in the form of $\{\mb{e}_A; \mb{e}_V\}$.
Then, a pairwise Euclidean distance matrix is obtained by computing the distance between all paired embeddings. Afterwards, for each embedding (either audio or video), another two embeddings are chosen from the same batch, to form a hard triplet. It is worth mentioning that, when generating the hardest positive or negative pair, we take both the intermodal and intramodal similarity into consideration. In this manner, the learning process {\blue forces} the model to narrow the distribution gap of embeddings from different modalities, and to keep the specific emotional semantics intact in the meantime.

Supervised by the crossmodal triplet loss $\mc{L}_{VAT}$, the model is forced to minimise the optimisation objective $\mc{J}(\theta)$, which can be formulated as:
\begin{equation}
\label{eq:triplet}
    \mc{J}(\theta)=\mc{L}_{mon}+\beta \cdot \mc{L}_{VAT},
\end{equation}
where $\mc{L}_{mon}$ denotes the conventional monomodal discriminative loss, which is, in our case, $\mc{L}_{V}$ for facial emotion recognition or $\mc{L}_{A}$ for speech.

\subsection{Crossmodal Emotion Embedding}
\label{subsec:method_cross}
In this subsection, we illustrate the proposed EmoBed framework, which integrates the triplet constraint (see Section~\ref{subsec:method_mono}) into the joint training approach (see Section~\ref{subsec:method_joint}). The overall training process is demonstrated in Fig.~\ref{fig:training}.

Generally, after extracting monomodal descriptors from standard pre-processing procedures, embedding functions $f_A(\cdot)$ and $f_V(\cdot)$ are estimated by two embedded neural networks, respectively, which project audio and video descriptors into a common latent space. Subsequently, the audio and visual embeddings are fed into a shared emotion recognition neural network, which is trained via a joint training loss. Concurrently, the training process is supervised by the triplet loss of the audio and visual embeddings.

Mathematically, when applying the EmoBed framework for audio emotion recognition, the objective function can be formatted as:
\begin{equation}
\label{eq:cross_audio}
    \mc{J}(\theta) = \mc{L}_{A} + \alpha \cdot \mc{L}_{V} + \beta \cdot \mc{L}_{VAT} + \lambda \cdot \mc{R}(\theta),
\end{equation}
where $\mc{L}_{A}$ and $\mc{L}_{V}$ represent the discriminative loss function of audio and visual data, respectively, while $\mc{L}_{VAT}$ represents the triplet loss function of both audio and visual data. Moreover, the hyperparameters $\alpha$ and $\beta$ are introduced to weight the contribution of the video data and the triplet loss. Furthermore, $\lambda$ is applied to control the importance of the regularisation term $\mc{R}(\theta)$ (L2).
Similarly, when training the EmoBed framework for facial emotion recognition, the objective function in Eq.~(\ref{eq:cross_audio}) can be modified by exchanging $\mc{L}_A$ and $\mc{L}_V$.


After the model has finished training, the components associated with the auxiliary modality can be discarded, while the rest is retained and utilised to recognise emotions. 
\section{Experimental Implementation}
\label{sec:ex}
To evaluate our approach comprehensively, we conducted extensive experiments on two multimodal emotional datasets for two tasks, respectively. Specifically, the RECOLA dataset was chosen for \textit{dimensional emotion regression}, whereas the OMG-Emotion dataset for \textit{categorical emotion classification}. 
In this section, we briefly introduce the two datasets. Then, we describe the experimental setup in detail for the sake of experiment replication, and also the evaluation metrics for a performance comparison.
\subsection{Evaluated Databases and Features}
\label{subsec:ex_database}

\subsubsection{RECOLA}
\label{subsubsec:ex_db_recola}
This database is widely used for audiovisual dimensional emotion recognition, and a standard database which was previously applied in the AVEC since 2015~\cite{Valstar16-AVEC, Ringeval18-AVEC}. 
It contains audiovisual recordings of spontaneous and natural interactions from 27 French-speaking participants in order to investigate socio-affective behaviours in the context of remote collaborative tasks. Moreover, time- and value-continuous dimensional emotion annotations in terms of \textit{arousal} and \textit{valence} are given with a constant frame rate of 40\,ms for the first five minutes of each recording, by averaging all six annotators and meanwhile taking the inter-evaluator agreement into consideration~\cite{Ringeval13-Introducing}. The dataset is further equally divided into three disjoint parts, by balancing the gender, age, and mother tongue of the participants. Therefore, each part consists of nine unique recordings, resulting in 67.5\,k segments in total for each part (training, development, or test).

When conducting experiments on RECOLA, we employed the same acoustic and visual features as the features utilised to compute the AVEC 2015 and 2016 baselines {\cite{Ringeval15-AV+EC,Valstar16-AVEC}}, for a fair comparison with other methods.
Specifically, as for the acoustic features, the extended Geneva Minimalistic Acoustic Parameter Set (\textit{eGeMAPS}~\cite{Eyben16-Geneva}) was extracted on all audio recordings by our open-source open\textsc{SMILE} toolkit~\cite{Eyben10-OTM}. This resulted in a set of 88 acoustic features per segment.

In relation to the visual features, we utilised both the \textit{appearance} and \textit{geometric} standard features of the AVEC challenges.
That is, we investigated handcrafted video features rather than learnt features from a pre-trained VGG-Face net as shown in {\blue Fig.~\ref{fig:training}}, as the inputs of visual embedding nets. This is for a fair comparison with other methods.
Similar to the acoustic features, the arithmetic mean and the standard derivation were computed over the sequential handcrafted visual features of each frame using a sliding window of 8\,s with a step size of 40\,ms.
This process led to 168 appearance and 632 geometric visual features.

\subsubsection{OMG-Emotion}
\label{subsubsec:ex_db_omg}
In our experiments, the One-Minute Gradual-Emotional~(OMG-Emotion) Behavior dataset~\cite{Barros18-OMG} was employed for categorical emotion classification. The OMG-Emotion dataset is composed of 567 emotional monologue videos collected from Youtube, with an average length of one minute. These videos were then divided into utterance-level clips, and annotated by at least five annotators~\cite{Barros18-OMG}. Seven categorical emotions were considered, \ie \textit{neutral, happiness, sadness, anger, surprise, fear,} and \textit{disgust}. Majority voting was then applied to compute the gold standard based on all annotations of the same segment.
Moreover, the dataset is split into training, development, and test sets, resulting in 2\,440, 617, and 2\,229 segments for each partition, respectively.
Note that, in this work, we performed experiments and reported performances only on the development set, as labels of the test set are not yet accessible.

To extract acoustic features on the OMG-Emotion dataset, we used the eGeMAPS feature set~\cite{Eyben16-Geneva}, resulting in 88 features for each utterance. For visual descriptors, firstly, a multi-task cascaded CNN~\cite{Zhang16-JFD} was applied for face detection and alignment on each frame. After that, frame-level intermediate deep features of size 4\,096 were extracted from the ``$fc$-7" layer of the VGG-Face model~\cite{Parkhi15-DFR}, which was pre-trained on a large number of facial images. Lastly, average pooling was conducted on all frames of the same clip to deliver the final utterance-level video descriptors.


\subsection{Experimental Setup}
\label{subsec:ex_setup}
The proposed EmoBed networks were implemented with GRU-RNNs. Compared with LSTM units, the employed GRUs have fewer parameters owing to the lack of separate memory cells and output gates, which results in a faster training process and a less-training-data demand for achieving a good generalisation~\cite{Cho14-properties}. Moreover, many empirical evaluations have shown that GRUs perform as competitively as LSTM units~\cite{Jozefowicz15-empirical}.

For the RECOLA experiments, we fixed the number of hidden layers for the modality-specific subnetworks (\ie for audio and video) and the modality-shared subnetwork  to be two, respectively. Each hidden layer has 120 hidden nodes. To train the network, we utilised the Adam optimisation algorithm with an initial learning rate of 0.001. Moreover, we employed a regularisation term (L2) with a weight decay of $10^{-4}$, to improve the model generality. Furthermore, to facilitate the training process, we set the mini-batch size to be four. Additionally, an online standardisation was always applied to the input data by using the means and variations of the training set. All these settings were empirically recommended by our previous work on the RECOLA database after a grid search evaluation strategy~\cite{Zhang18-Dynamic}. 

For the OMG-Emotion experiments, we kept in line with the network and the training hyper-parameters, but used one hidden layer as the modality-specific or the modality-shared subnetworks due to the limited size of the OMG-Emotion dataset, and used 64 as the mini-batch size. 

When training the network in a crossmodal scenario, we randomly chose the audio and video data, rather than the aligned data pair across audio and video, as the mini-batch. The advantage of this method is that, it does not require the synchronous presence of both modalities in the training phase. This means that we can principally mix the audio-only and video-only databases to complete the network training process. 

Additionally, when continuously assessing emotions, the annotators have to take time to perceive acoustic events, understand them, and report the emotional states~\cite{Mariooryad15-Correcting}. To address this annotation delay problem, we took a widely used explicit compensation approach. That is, we shifted the gold standard back in time (\ie 2.4\,s) with respect to the features for all modalities and tasks~\cite{Valstar16-AVEC}, with an assumption that the delay is invariant with annotators, annotator states, modalities, and tasks. 

To refine the obtained predictions for continuous emotion recognition, we further performed a chain of post-processing, including median filtering, centring, scaling, and time-shifting, as suggested {\blue by Valstar~\et~\cite{Valstar16-AVEC}}.
The filtering window size $W$ (from 0.12\,s to 0.44\,s at a rate of 0.08\,s) and the time-shifting delay $D$ (from 0.04\,s to 0.60\,s at a step of 0.04\,s) were optimised using a grid search method. All the post-processing parameters were optimised on the development set and then applied to the test set. 

\subsection{Evaluation Metrics}
\label{subsec:ex_metrics}

To evaluate the performance of the continuous emotion regression model, we took the {\em Concordance Correlation Coefficient} (CCC), which was officially recommended by the AVEC 2015-2018 challenges~\cite{Valstar16-AVEC}. The CCC is defined by:
  \begin{equation}
  r_c = \frac{2r\sigma_x\sigma_y}{\sigma^2_x+\sigma^2_y+(\mu_x - \mu_y)^2},
  \label{eq5}
  \end{equation}
where $r$ represents {\em Pearson's correlation coefficient} {\blue (PCC)} between two time series (\eg prediction and gold-standard), $\mu_x$ and $\mu_y$ denote the mean of each time series, and $\sigma^2_x$ and $\sigma^2_y$ stand for the corresponding variances. 
Compared with PCC, the CCC considers not only the shape similarity between the two series but also the value precision. This is especially relevant for estimating the performance of time-continuous emotion prediction models, as both the trends as well as absolute prediction values are relevant for describing the performance of a model. The CCC metric falls into the range of [-1, 1], where $+$1 represents perfect concordance, $-$1 total discordance, and 0 no concordance at all. 
  
As to the discrete emotion classification tasks, we chose F1 as the evaluation metric, mainly due to the {\blue fact} that i) it can provide an overview performance in a multi-class setting as it is calculated by the harmonic mean of unweighted precision and recall; ii) it was employed by the OMG-Emotion challenge in 2018~\cite{Barros18-OMG}. In general, a higher CCC or F1 indicates a better prediction performance.

Further, to evaluate the statistical significance of performance improvement, unless stated otherwise, we undertook a \textit{Fisher r-to-z transformation}~\cite{Cohen13-Applied} for dimensional continuous emotion regression and a one-tailed $z$-test for categorical discrete emotion classification. Only if the $p$-value was lower than .05, a significant difference was triggered.

\section{Experimental Results and Discussions}
\label{sec:res}
For the sake of fair performance comparison, in this section, we report on conducted experiments on two emotional databases (\ie {\blue RECOLA} and OMG-Emotion) with their corresponding standard testbeds of the AVEC 2016~\cite{Valstar16-AVEC} and OMG-Emotion challenges 2018~\cite{Barros18-OMG}.  

\subsection{Results on RECOLA}
\label{subsec:res_recola}

\begin{table*}[!ht]
    \centering
    \caption{\blue Performance comparison in terms of CCC for the {\bf arousal} prediction among the proposed EmoBed systems, related baselines, and other state-of-the-art systems. 
    These results pertain to the experiments conducted on the \textit{dev}elopment and \textit{test} partitions of the RECOLA database. Three feature sets (audio-eGeMAPS, video-appearance, and video-geometric) were employed to evaluate all approaches. Four monomodal scenarios are considered: audio (+video-app.), video-app. (+audio), audio (+video-geo.), video-geo. (+audio), where the modalities in the parentheses are the employed auxiliary modalities. The cases where EmoBed systems have a statistical significance of performance improvement over the classic monomodal systems are marked by the ``$\star$'' symbol.  MTL: multi-talk learning; DDAT: dynamic difficulty awareness training; RE: reconstruction error; PU: perception uncertainty.}
    \begin{threeparttable}
    \begin{tabular}{lcccccccccccc}
    \toprule
    \multirow{2}{*}{\quad CCC}	     && \multicolumn{2}{c}{audio (+video-app.)} && \multicolumn{2}{c}{video-app. (+audio)} && \multicolumn{2}{c}{audio (+video-geo.)} && \multicolumn{2}{c}{video-geo. (+audio)}  \\
		\cmidrule{3-4} \cmidrule{6-7} \cmidrule{9-10} \cmidrule{12-13}
	\em{our frameworks}	&& dev   & test  && dev & test  && dev   & test && dev   & test \\

    \midrule
    \quad classic monomodal 				        && .766 & .605 && .512 & .411 && .766 & .605 && .499 & .399 \\
    \quad joint audiovisual training 	&& .769 & .611 && .520 & .401 && .769 & .611 && .515 & .413 \\
    \quad crossmodal triplet training 			    && {.795} & .633 && .541 & .465 && {.794} & .632 && .512 & .397 \\
    \quad EmoBed 		    && .792$^\star$ & {.644}$^\star$ && \bf{.549}$^\star$ & {\bf .475}$^\star$ && .793$^\star$ & {.639}$^\star$ && {.527}$^\star$ & {\bf .417}$^\star$ \\

    \midrule
    \multirow{2}{*}{}	     && \multicolumn{2}{c}{audio-only} && \multicolumn{2}{c}{video-app.-only} && \multicolumn{2}{c}{audio-only} && \multicolumn{2}{c}{video-geo.-only}  \\
    	\cmidrule{3-4} \cmidrule{6-7} \cmidrule{9-10} \cmidrule{12-13}
    \em{state of the art}	&& dev   & test  && dev & test  && dev   & test && dev   & test \\
    \midrule
    \quad SVR + offset, 2016~\cite{Valstar16-AVEC} 	&  & .796 & .648 &  & .483 & .343 &&.796 & .648&  & .379 & .272 \\
    \quad Strength modelling, 2017~\cite{Han17-Strength} 	&  & .755 & .666 &  & .350  & .196 &&.755 & .666&  & ---     & ---     \\
    \quad End-to-end, 2017~\cite{Tzirakis17-End} 		&  & .786 & \bf{.715} &  & .371 & .435 &&.786 & \bf{.715}&  & ---     & ---     \\
    \quad MTL (RE), 2017~\cite{Han17-Reconstruction} 		&  & .788 & .629 &  & .512 & .425 &&.788 & .629&  & .502 & .324 \\ 
    \quad MTL (PU), 2017~\cite{Han17-From} 			&  & .803 & .654 &  & .502 & .406 &&.803 & .654&  & .508 & .327 \\
    \quad Curriculum learning, 2018~\cite{Lotfian18-Curriculum}		&  & .687 & .591 &  & .417 & .343 &&.687 & .591&  & .394 & .267 \\
    \quad DDAT (RE), 2018~\cite{Zhang18-Dynamic}		&  & .807 & .694 &  & .539 & .437 &&.807 & .694&  & \bf{.544} & .400   \\
    \quad DDAT (PU), 2018~\cite{Zhang18-Dynamic}			&  & \bf{.811} & .664 &  & .518 & .438 &&\bf{.811} & .664&  & .513 & .397 \\
    \bottomrule
    \end{tabular}
    \end{threeparttable}
    \label{tab:res_recola_arousal}
\end{table*}

\begin{table*}[!ht]
    \centering
    \caption{\blue Performance comparison in terms of CCC for the {\bf valence} prediction among the proposed EmoBed systems, related baselines, and other state-of-the-art systems. 
    These results pertain to the experiments conducted on the \textit{dev}elopment and \textit{test} partitions of the RECOLA database. Three feature sets (audio-eGeMAPS, video-appearance, and video-geometric) were employed to evaluate all approaches. Four monomodal scenarios are considered: audio (+video-app.), video-app. (+audio), audio (+video-geo.), video-geo. (+audio), where the modalities in the parentheses are the employed auxiliary modalities. The cases where EmoBed systems have a statistical significance of performance improvement over the classic monomodal systems are marked by the ``$\star$'' symbol.  MTL: multi-talk learning; DDAT: dynamic difficulty awareness training; RE: reconstruction error; PU: perception uncertainty.}
    \begin{threeparttable}
    \begin{tabular}{lcccccccccccc}
    \toprule
    \multirow{2}{*}{\quad CCC}	     && \multicolumn{2}{c}{audio (+video-app.)} && \multicolumn{2}{c}{video-app. (+audio)} && \multicolumn{2}{c}{audio (+video-geo.)} && \multicolumn{2}{c}{video-geo. (+audio)}  \\
		\cmidrule{3-4} \cmidrule{6-7} \cmidrule{9-10} \cmidrule{12-13}
	\em{our frameworks}	&& dev   & test  && dev & test  && dev   & test && dev   & test \\
		\midrule
    \quad classic monomodal		                &  & .504 & .381 &  & .545 & .525 &  & .504 & .381 &  & .619 & {\bf .529} \\
    \quad joint audiovisual training    &  & .505 & .393 &  & .546 & .511 &  & .513 & .395 &  & .622 & .527 \\
    \quad crossmodal triplet training              &  & {.515} & .404 &  & {.564} & {.517} &  & .512 & .405 &  & .636 & .517 \\
    \quad EmoBed             &  & .514$^\star$ & {\bf .434}$^\star$ &  & {.564}$^\star$ & .516 &  & {\bf .521}$^\star$ & {\bf .439}$^\star$ &  & {\bf .645}$^\star$ & .520 \\ 
    \midrule
    \multirow{3}{*}{}	     && \multicolumn{2}{c}{audio-only} && \multicolumn{2}{c}{video-app.-only} && \multicolumn{2}{c}{audio-only} && \multicolumn{2}{c}{video-geo.-only}  \\
    	\cmidrule{3-4} \cmidrule{6-7} \cmidrule{9-10} \cmidrule{12-13}
    \em{state of the art}	&& dev   & test  && dev & test  && dev   & test && dev   & test \\
    \midrule
    \quad SVR + offset, 2016~\cite{Valstar16-AVEC}  &  & .455 & .375 &  & .474 & .486 &  &.455 & .375&  & .612 & .507 \\
    \quad Strength modelling, 2017~\cite{Han17-Strength}  &  & .476 & .364 &  & .592 & .464 &  &.476 & .364&  & ---     & ---     \\
    \quad End-to-end, 2017~\cite{Tzirakis17-End}   &  & .428 & .369 &  & \bf{.637} & \bf{.620} &  &.428 & .369&  & ---     & ---     \\
    \quad MTL (RE), 2017~\cite{Han17-Reconstruction}  &  & \bf{.519} & .331 &  & .529 & .366 &  &.519 & .331  && .632 & .488 \\ 
    \quad MTL (PU), 2017~\cite{Han17-From}  &  & .506 & .416 &  & .468 & .418 &  &.506 & .416&  & .643 & .452 \\
    \quad Curriculum learning, 2018~\cite{Lotfian18-Curriculum}  &  & .159 & .174 &  & .446 & .419 &  &.159 & .174&  & .300 & .269 \\
    \quad DDAT (RE), 2018~\cite{Zhang18-Dynamic}  &  & .508 & .422 &  & .528 & .457 &  &.508 & .422&  & .639 & .471 \\
    \quad DDAT (PU), 2018~\cite{Zhang18-Dynamic}  &  & .498 & .407 &  & .514 & .431 &  & .498 & .407&  & .632 & .501 \\
    \bottomrule
    \end{tabular}
    \end{threeparttable}
    \label{tab:res_recola_valence}
\end{table*}

 
 {\blue As suggested in~\cite{Valstar16-AVEC}, for these experiments, we took two visual feature sets (\ie appearance and geometric) and one acoustic feature set (\ie eGeMAPS), as aforementioned in Section~\ref{subsubsec:ex_db_recola}. This leads to four experimental scenarios: audio (+video-app.), video-app. (+audio), audio (+video-geo.), and video-geo. (+audio), where the modalities shown in the parentheses indicate the employed auxiliary modalities for the corresponding single modalities.} In addition, we evaluated the systems on both dimensional {\em arousal} and {\em valence} regressions as well. 
For the classic monomodal systems, we independently performed the training process on either audio or video data. That was achieved by setting $\alpha$ to be 0.0 in Eq.~(\ref{eq:fuse1}) and Eq.~(\ref{eq:fuse2}), respectively. 
 
The obtained results for arousal and valence predictions are presented in Tables~\ref{tab:res_recola_arousal} and~\ref{tab:res_recola_valence}, respectively.  
From the two tables, it can be seen that our classic monomodal systems outperform the challenge benchmarks that used the `SVR + offset' approach~\cite{Valstar16-AVEC} in most cases. One exception is the arousal prediction with audio signals, which probably {\blue is attributed} to the fact that a fixed network structure, rather than the optimised one on the arousal prediction, was employed in our experiments (see Section~\ref{subsec:ex_setup}). These results further confirm that GRU-RNNs hold the powerful capability to capture {\blue long-term} context dependence. 

Furthermore, when jointly training audio and video data (see Section~\ref{subsec:method_joint}), one may notice that the corresponding monomodal systems (based on either audio-eGeMAPS, video-appearance, or video-geometric) can deliver higher CCCs compared with the classic monomodal systems, in seven out of eight cases for the arousal prediction and six out of eight cases for the valence prediction, respectively. This observation implies that such a joint training process can somewhat transfer shared semantic information from other heterogeneous data to the target modality thanks to the implementations of i) a shared subnetwork and ii) a multi-task learning framework. 

Rather than the joint audiovisual training, when performing the triplet constraint across the audio and video modalities (see Section~\ref{subsec:method_mono}), the obtained CCCs (the third lines in Tables~\ref{tab:res_recola_arousal} and~\ref{tab:res_recola_valence}) show that the introduced enhanced monomodal systems  again generally significantly outperform the classic monomodal systems in most cases (Fisher $r$-to-$z$ transformation, $p<$.05). 
This suggests that the implementation of the triplet constraint is helpful to distil emotional discriminative representations, not only in a monomodal scenario~\cite{Han18-Emotion} but also in a crossmodal scenario as investigated in this article. 

Finally, when we simultaneously carried out the crossmodal triplet training as well as the joint audiovisual training processes, it can be seen that the EmoBed systems achieve the best performance in most cases, \ie six out of eight cases for the arousal regression and five out of eight cases for the valence regression. For example, the obtained CCCs on the test set of the audio-based model are boosted from .605 to .644 and .639 for arousal regression, and from .381 to .434 and .439 for valence regression, when respectively integrating with video-appearance and video-geometric feature sets in the training process. Furthermore, the best CCCs achieved by the video-based models reach to .475 and .417 with appearance and geometric feature sets, respectively, with absolute CCC increases of .064 and .018 compared with the classic monomodal systems for arousal regression. 
{\blue Such an observation, however, cannot be found for the video-based valence regression models on the test set. This exception possibly is attributed to the distribution mismatch between the development and test partitions. 
Overall, it is concluded that the proposed EmoBed can largely supply additional knowledge from audio signals to alleviate the shortage of video signals, and vice verse.}

Meanwhile, as presented in Tables~\ref{tab:res_recola_arousal} and~\ref{tab:res_recola_valence}, the EmoBed systems achieve comparable or better performance to other state-of-the-art methods, such as the 
end-to-end systems~\cite{Tzirakis17-End}, curriculum learning systems~\cite{Lotfian18-Curriculum}, multi-task learning systems~\cite{Han17-From}, and Dynamic Difficulty Awareness Training (DDAT) systems~\cite{Zhang18-Dynamic}. 
{\blue Although most of these reported systems utilised a variety of multimodal fusion approaches to boost their final performance, these systems, in the reference stage, demand the simultaneous occurrence of all modalities that appear in the training stage, to the best of our knowledge.
It also has to be noted that the best achieved results are delivered by using the end-to-end system for the arousal predictions when using audio signals (\ie .715 of CCC) and for the valence predictions when using video signals (\ie .620 of CCC). This observation further confirms that optimising the whole network from the raw signals to the target can benefit the efficient extraction of high-level representations. }
{\blue Therefore, in the future we will further implement the EmoBed system in an end-to-end fashion, which is expected to further enhance the performance of monomodal emotion recognition.}

\subsection{Results on OMG-Emotion} 
\label{subsec:res_omg}

\begin{table}[!t]
    \centering
    \caption{Performance of the proposed EmoBed systems, related baselines, and other reported systems in terms of F1 on the development set of the OMG-Emotion dataset. The cases where EmoBed systems have a statistical significance of performance improvement over the classic monomodal systems are marked by the ``$\star$'' symbol.}
    \begin{threeparttable}
    \begin{tabular}{llcl}
    \toprule
    \multirow{1}{*}{\quad F1 [\%]}	     & \multicolumn{1}{c}{audio} && \multicolumn{1}{c}{video}  \\
    \midrule
    {\em other approaches} \\
    \quad SVM~\cite{Barros18-OMG} & 33.0 && --- \\
    \quad RF~\cite{Barros18-OMG} & --- && 37.0 \\
    \midrule
    {\em our frameworks} \\ 
    \quad classic monomodal & 36.5 && 37.9\\ 
    \quad joint audiovisual training & 40.2 && 42.1 \\
    \quad crossmodal triplet training & 40.7 && 41.0 \\
    \quad EmoBed & {\bf 41.7}$^\star$ && {\bf 43.9}$^\star$ \\
    \bottomrule
    \end{tabular}
    \end{threeparttable}
    \label{tab:res_omg}
\end{table}

For our experiments on the OMG-Emotion database, we conducted seven-class categorical emotion classification tasks on audio and visual signals. Table~\ref{tab:res_omg} presents the performance of the models in terms of F1 on the development set (note that the annotations of the test set are not publicly available).
From the table, we can see that on this database, our classic monomodal models outperform the other methods reported in the literature~\cite{Barros18-OMG}, \ie Support Vector Machine (SVM) and Random Forest (RF). More specifically, our classic monomodal models yields higher F1 than SVM (36.5\,\% vs 33.0\,\%) for audio, and than RF (37.9\,\% vs 37.0\,\%) for video. 

Additionally, comparing the proposed joint audiovisual training models with the classic monomodal systems, it is noticed that the former approach outperforms the latter one by a large margin, \ie 40.2\,\% vs 36.5\,\% for audio and 42.1\,\% vs 37.9\,\% for video.
These experimental results again indicate that, the proposed joint audiovisual training approach is plausible to promote performances of monomodal emotion classification.
Furthermore, similar results are also obtained when utilising the triplet training approach to distil the salient representations across multiple modalities. 
Nevertheless, the highest F1s are achieved by means of the EmoBed systems, which deliver 5.2\,\% and 6.0\,\% absolute performance gain compared with the classic monomodal systems when using audio or video signals, respectively. 
All these observations further confirm the findings discovered from the RECOLA database.

\subsection{Visualisation of Emotion Embeddings}
\label{subsec:res_vis}

\begin{figure}[t]
    \centering
    \subfigure[arousal (classic monomodal)]{
    \includegraphics[height=1.6in,width=1.65in, trim={1.cm 0.6cm 1.6cm 1cm}, clip]{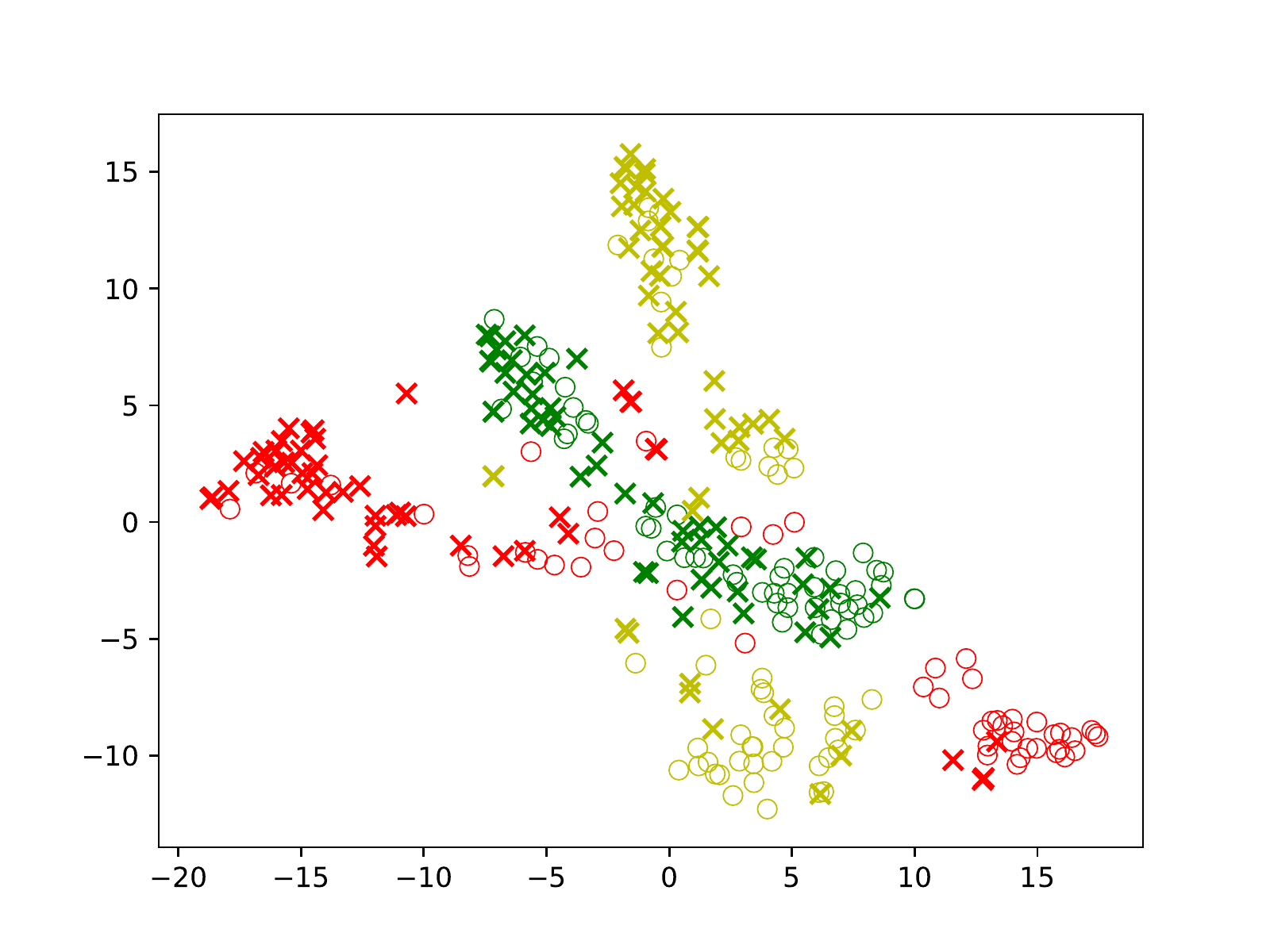}
    }
    \subfigure[arousal (EmoBed)]{
    \includegraphics[height=1.6in, width=1.65in,trim={1.cm 0.6cm 1.6cm 1cm}, clip]{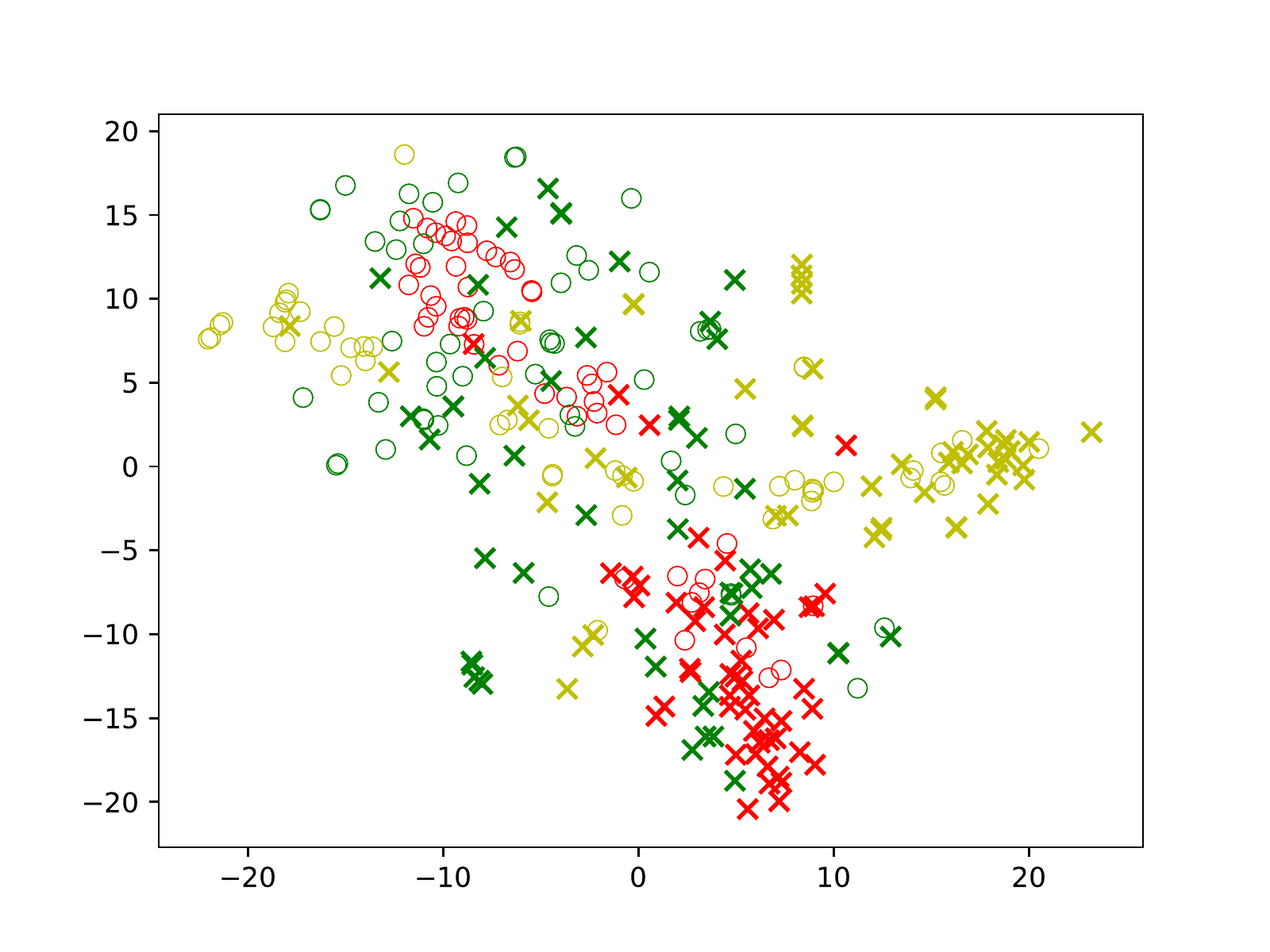}
    }
    \subfigure[valence (classic monomodal)]{
    \includegraphics[height=1.6in,width=1.65in, trim={1.cm 0.6cm 1.6cm 1cm}, clip]{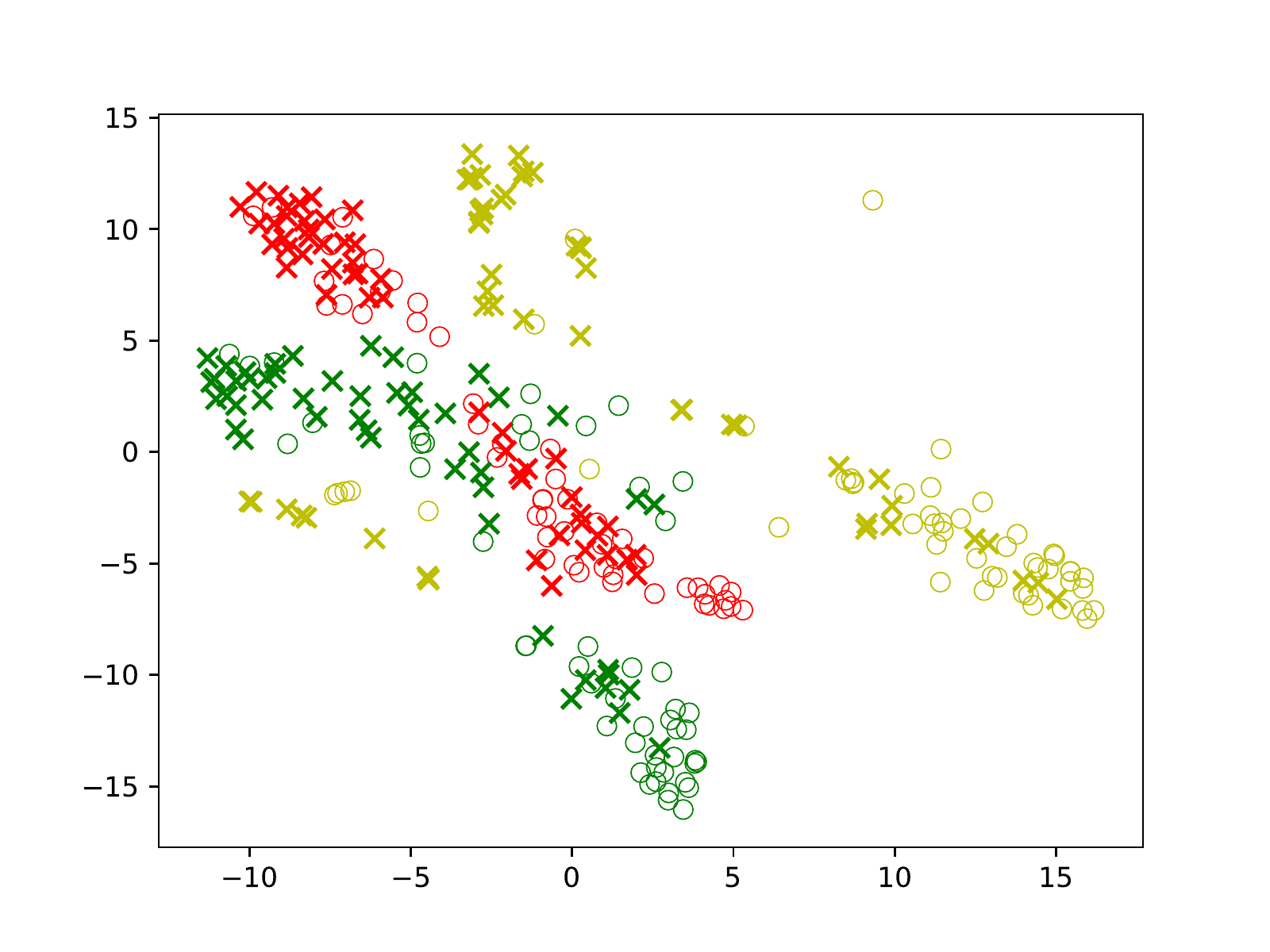}
    }
    \subfigure[valence (EmoBed)]{
    \includegraphics[height=1.6in, width=1.65in,trim={1.cm 0.6cm 1.6cm 1cm}, clip]{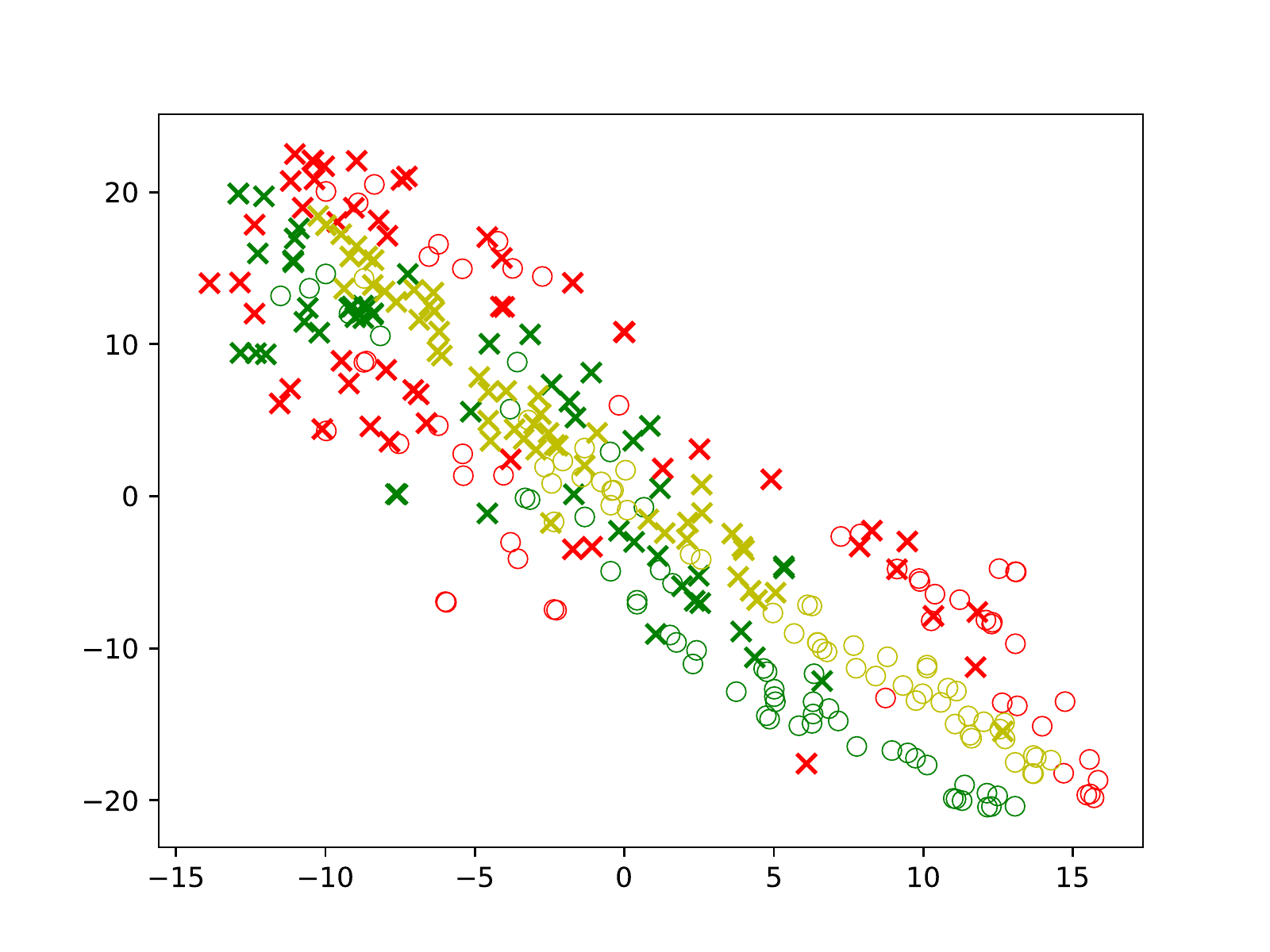}
    }
    
    \caption{\blue Visualisation of the learnt representations of the development set of the {\bf RECOLA} database when using the proposed EmoBed systems or the classic monomodal systems. Red, green, and yellow markers: representations from audio (eGeMAPS), video (appearance), and video (geometric) modalities; circle and cross markers: high and low arousal/valence.}
    \label{fig:res_tsne_recola}
\end{figure}

To investigate how the proposed crossmodal learning framework {\blue benefit} emotion recognition, we extracted the learnt representations from the classic monomodal systems and the proposed EmoBed systems. 
Fig.~\ref{fig:res_tsne_recola} illustrates the distribution of the learnt representations on the development set of the RECOLA database by means of t-Distributed Stochastic Neighbour Embedding (t-SNE). {\blue It} can be seen that with the classic monomodal systems, the learnt representations can be easily distinguished into three parts by the modalities they stem from, in either arousal (cf.~Fig.~\ref{fig:res_tsne_recola} (a)) or valence (cf.~Fig.~\ref{fig:res_tsne_recola} (c)) prediction. Specifically, the representations learnt from different modalities almost have no overlap albeit they belong to the same emotion states.  
In stark contrast, the representations extracted from EmoBed systems are visibly clustered together based on their emotional properties (cf.~Fig.~\ref{fig:res_tsne_recola} (b) and (d) for arousal and valence, respectively). 

\begin{figure}[!t]
    \centering
    \subfigure[classic monomodal]{
    \includegraphics[height=1.6in,width=1.65in, trim={1.cm 0.6cm 1.6cm 1cm}, clip]{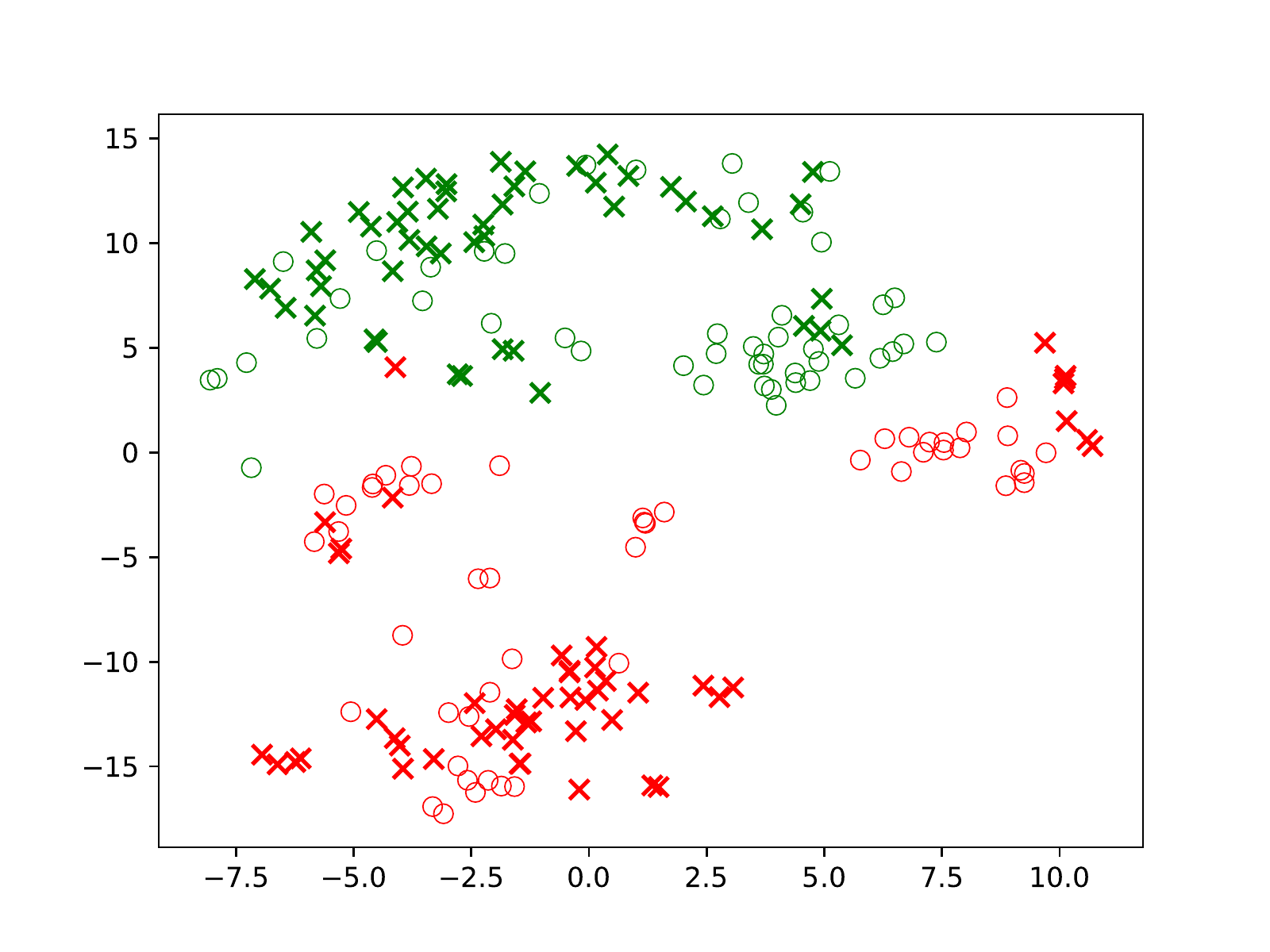}
    }
    \subfigure[EmoBed]{
    \includegraphics[height=1.6in, width=1.65in,trim={1.cm 0.6cm 1.6cm 1cm}, clip]{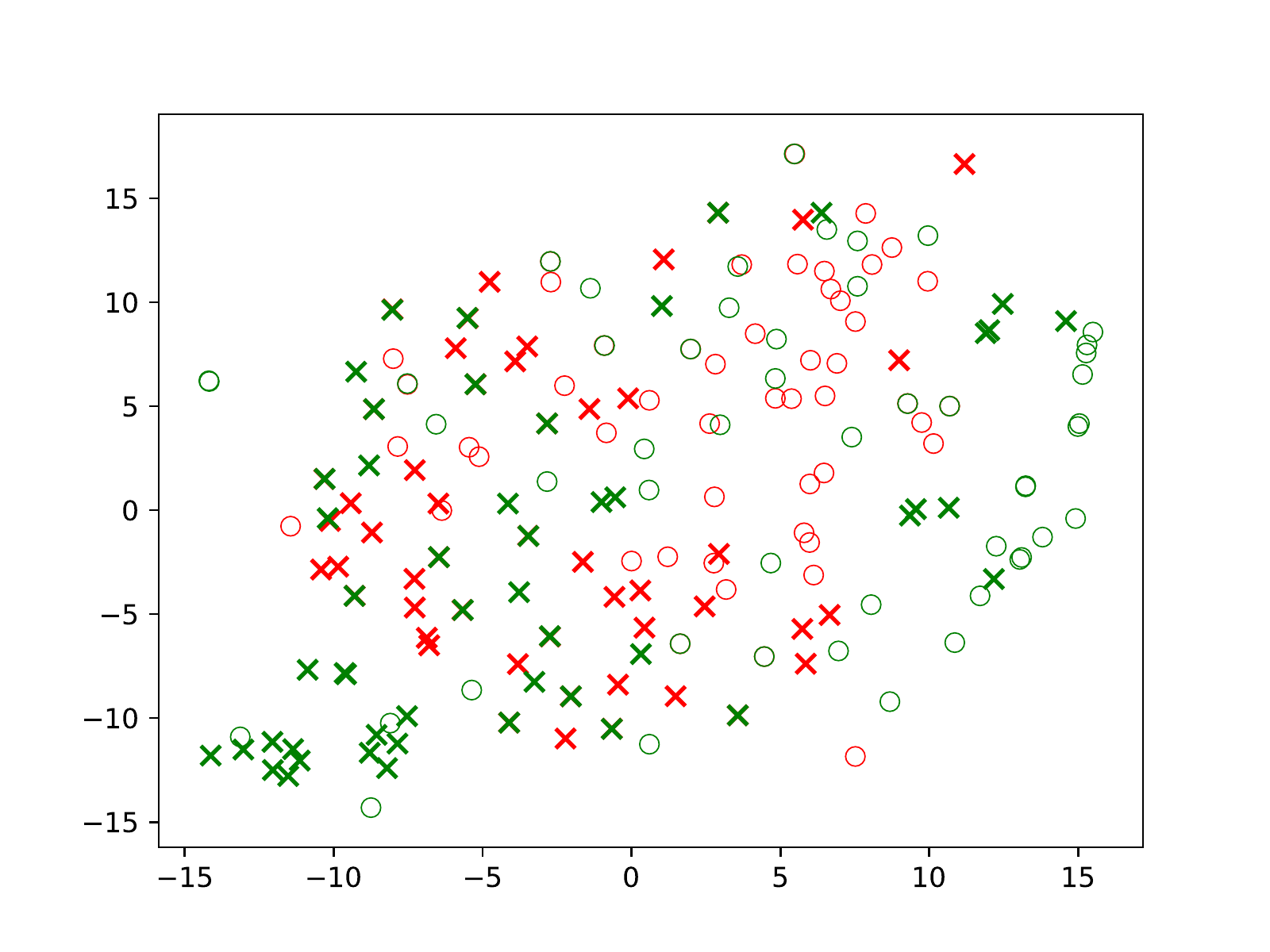}
    }
    \caption{\blue Visualisation of the learnt representations of the development set of the {\bf OMG-Emotion} database when using the proposed EmoBed systems or the classic monomodal systems. Red and green markers: representations from audio and video modalities; circle and cross markers: neutral and sad categories.}
    \label{fig:res_tsne_omg}
\end{figure}

Such an observation is even more noticeable on the OMG-Emotion database (cf.~Fig.~\ref{fig:res_tsne_omg}). Note that, for the sake of simplicity, we merely chose two emotional categories (\ie sad and neutral) for visualisation. Likewise, one can find that the representations belonging to the same emotional category share almost the same latent space. 

These findings indicate that the representations learnt by the proposed EmoBed are somewhat invariant to the modalities. By making use of the emotion embedding space, the emotional representations extracted from audio and video signals are able to implicitly fuse the knowledge from each other. Thus, the exploitation of mutual information possibly leads to performance improvement for a monomodal system. 

\subsection{Discussion} 
\label{subsec:res_dis}

To demonstrate the importance of the learning from auxiliary modalities for the monomodal emotion recognition system, we independently investigated the impact of weight change with respect to the counterpart modality, 
when in a joint audiovisual training process or in a crossmodal triplet training process. 

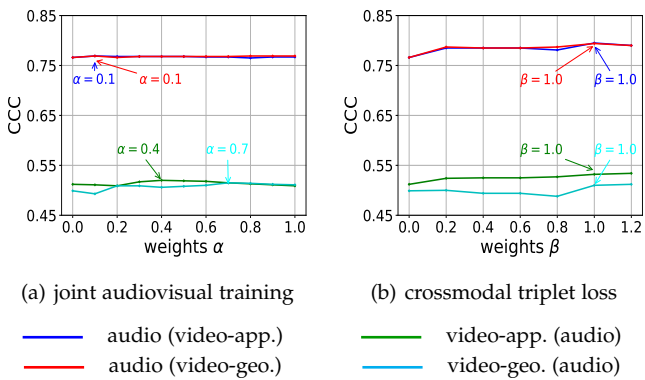
\begin{figure}[!t]
    \centering
    \subfigure[joint audiovisual training]{
    \includegraphics[height=1.4in,width=1.65in, trim={0cm 0cm 0cm 0cm}, clip]{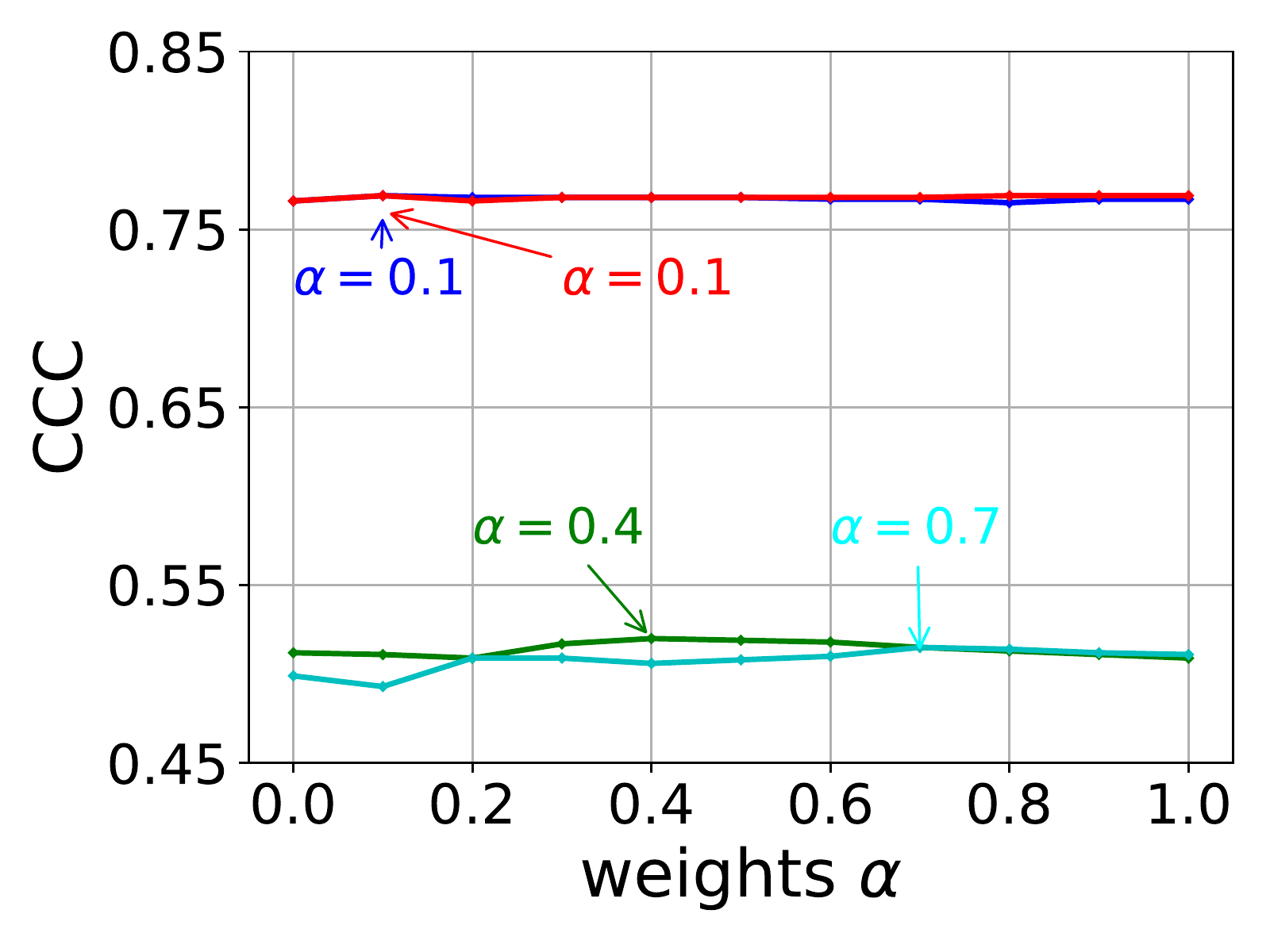}
    }
    \subfigure[crossmodal triplet loss]{
    \includegraphics[height=1.4in,width=1.65in, trim={0cm 0cm 0cm 0cm}, clip]{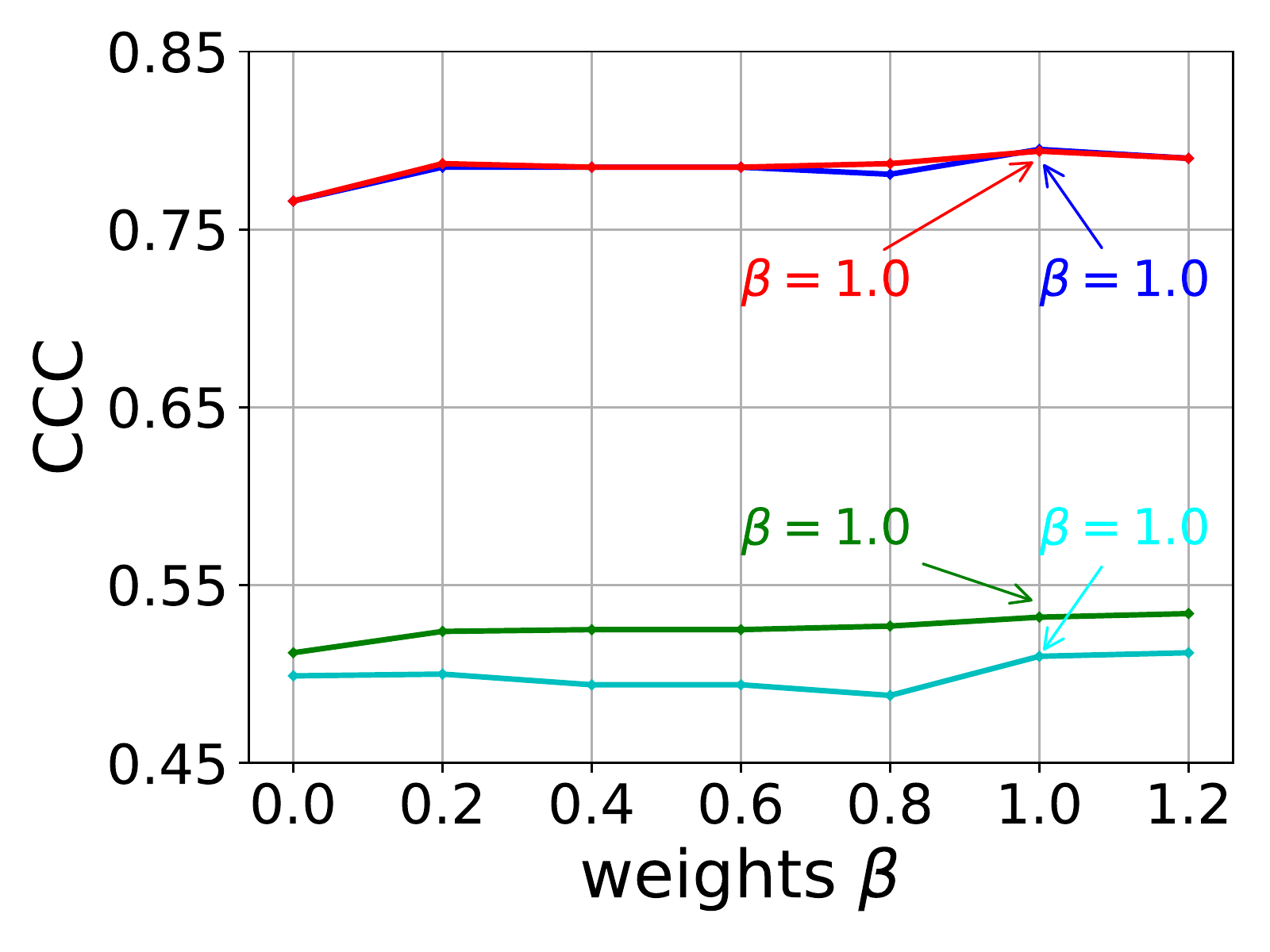}
    }
%
%
%
%
%

   \def\hsep{4cm}
   \begin{tikzpicture}[shorten >=0pt, draw=black!60, node distance=\hsep, line width=1pt, font=\footnotesize]
  
      \tikzstyle{symbol} = [text centered, text width=1cm, font=\footnotesize]; 
      \tikzstyle{node} = [font=\footnotesize]; 

      \node[font=\footnotesize] (label1) at (2,0) {audio (video-app.)};
      \node[] (label2) at (6.5,0) {video-app. (audio)};
      \node[] (label3) at (2,-.4) {audio (video-geo.)};
      \node[] (label4) at (6.5,-.4) {video-geo. (audio)};

      \draw [color=blue, ] ($(label1.west) + (-1,0)$)  --  ($(label1.west) + (-0.2,0)$); 
      \draw [color=black!40!green, ] ($(label2.west) + (-1,0)$)  --  ($(label2.west) + (-0.2,0)$); 
      \draw [color=red, ] ($(label3.west) + (-1,0)$)  --  ($(label3.west) + (-0.2,0)$); 
      \draw [color=cyan, ] ($(label4.west) + (-1,0)$)  --  ($(label4.west) + (-0.2,0)$); 
      
   \end{tikzpicture}

    \vspace{-.2cm}
    \caption{\blue Impact of the joint auxiliary modality loss on the {\em joint audiovisual training} systems (a), and impact of the crossmodal triplet loss on the {\em crossmodal triplet training} systems (b), with the {\bf RECOLA} database for {\bf arousal} regression. The best performed $\alpha$ or $\beta$ is indicated in each case.}
    \label{fig:res_weights_recola_joint}
\end{figure}

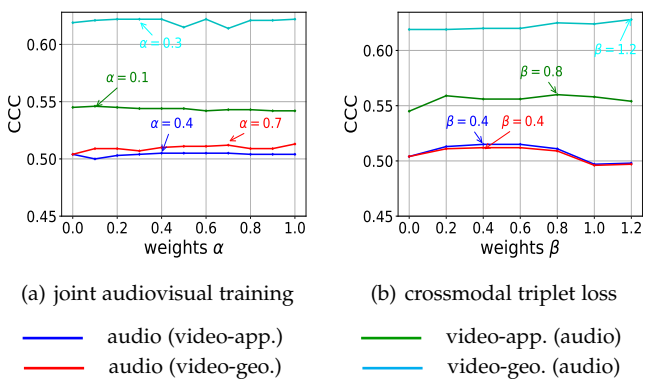
\begin{figure}[!t]
    \centering
    
    \subfigure[joint audiovisual training]{
    \includegraphics[height=1.4in, width=1.65in,trim={0cm 0cm 0cm 0cm}, clip]{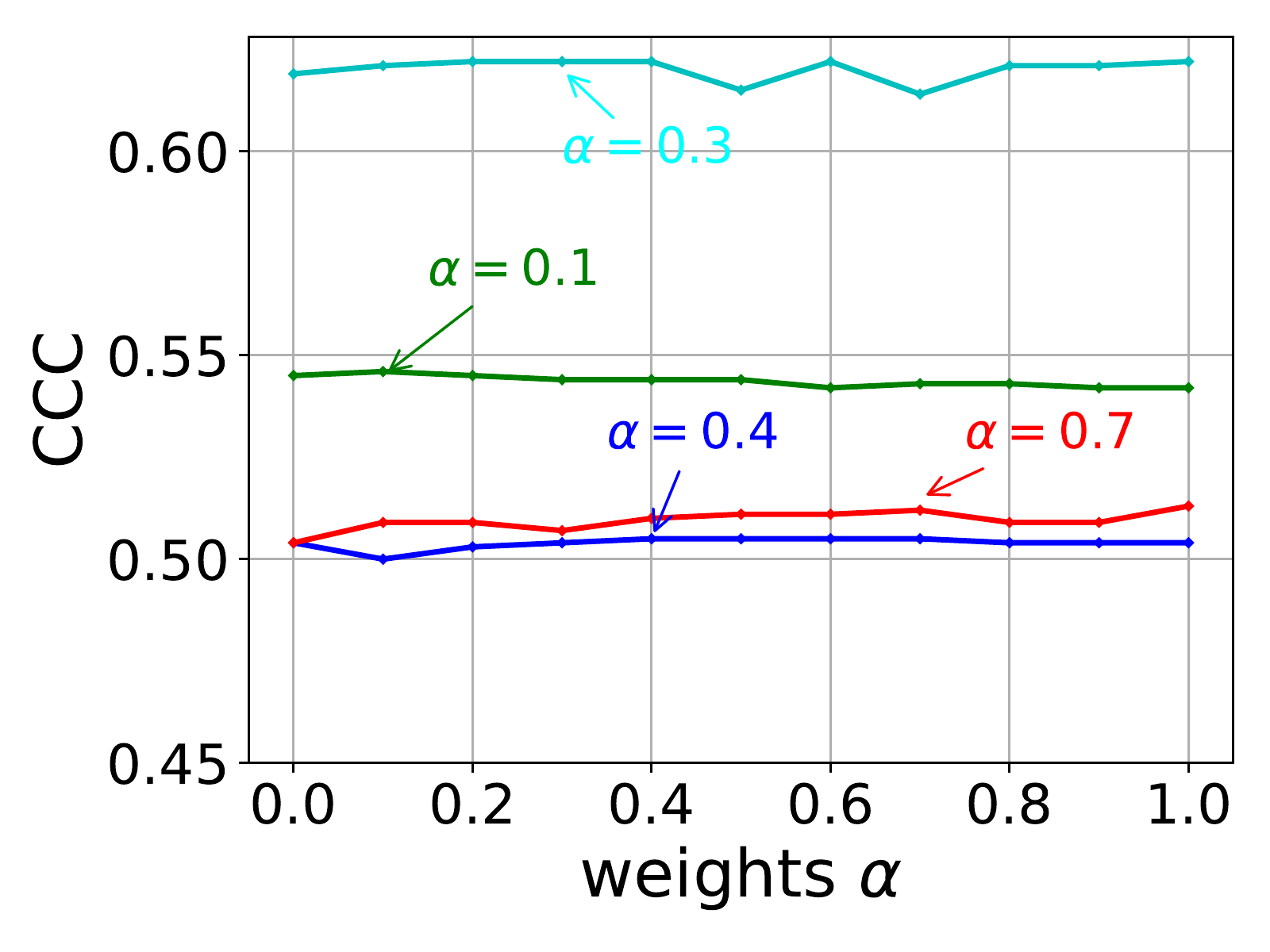}
    }
    \subfigure[crossmodal triplet loss]{
    \includegraphics[height=1.4in, width=1.65in,trim={0cm 0cm 0cm 0cm}, clip]{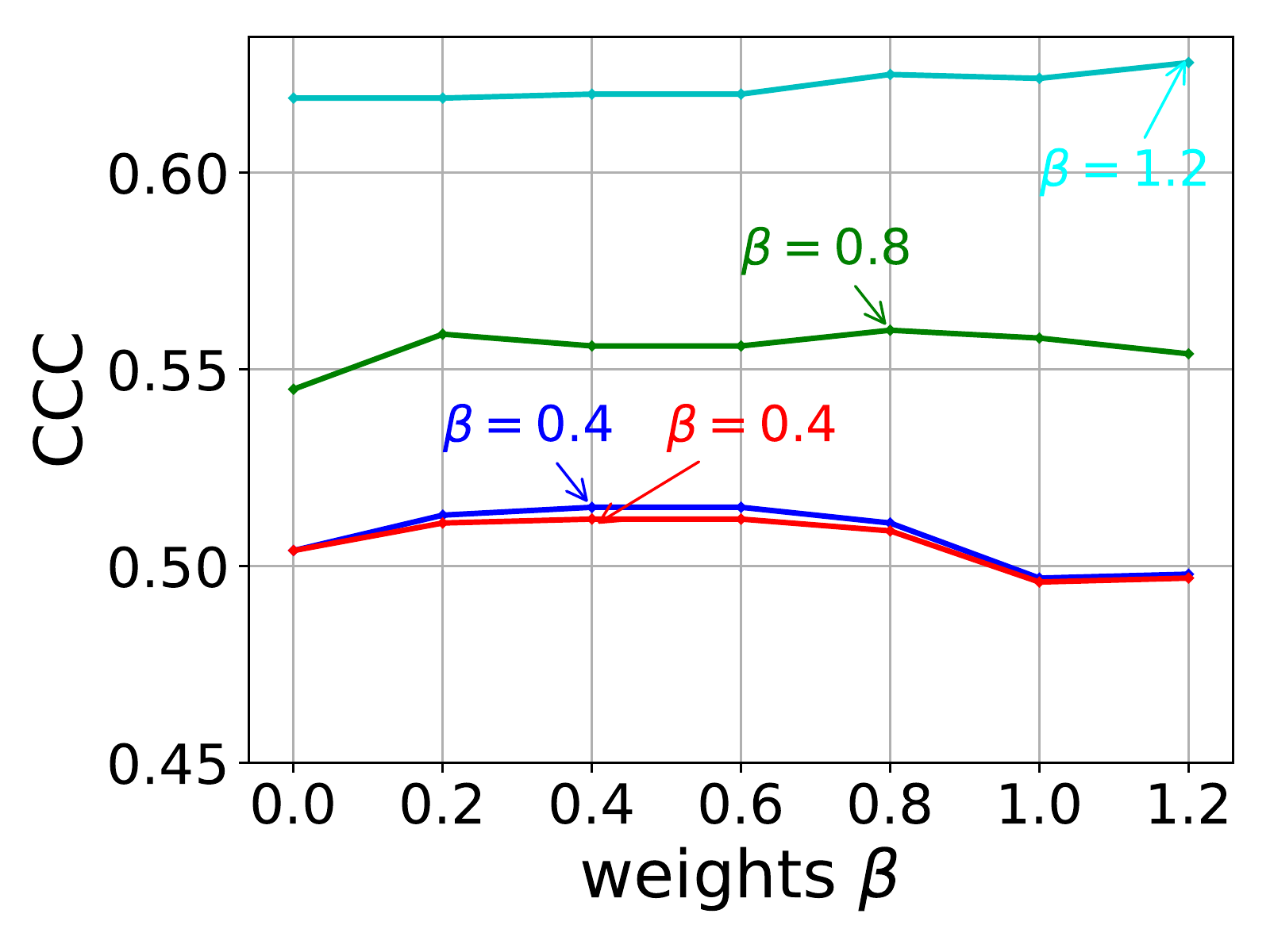}
    }
%
%
%
%
%

   \def\hsep{4cm}
   \begin{tikzpicture}[shorten >=0pt, draw=black!60, node distance=\hsep, line width=1pt, font=\footnotesize]
  
      \tikzstyle{symbol} = [text centered, text width=1cm, font=\footnotesize]; 
      \tikzstyle{node} = [font=\footnotesize]; 

      \node[font=\footnotesize] (label1) at (2,0) {audio (video-app.)};
      \node[] (label2) at (6.5,0) {video-app. (audio)};
      \node[] (label3) at (2,-.4) {audio (video-geo.)};
      \node[] (label4) at (6.5,-.4) {video-geo. (audio)};

      \draw [color=blue, ] ($(label1.west) + (-1,0)$)  --  ($(label1.west) + (-0.2,0)$); 
      \draw [color=black!40!green, ] ($(label2.west) + (-1,0)$)  --  ($(label2.west) + (-0.2,0)$); 
      \draw [color=red, ] ($(label3.west) + (-1,0)$)  --  ($(label3.west) + (-0.2,0)$); 
      \draw [color=cyan, ] ($(label4.west) + (-1,0)$)  --  ($(label4.west) + (-0.2,0)$); 
      
   \end{tikzpicture}

    \vspace{-.2cm}
    \caption{\blue Impact of the joint auxiliary modality loss on the {\em joint audiovisual training} systems (a), and impact of the crossmodal triplet loss on the {\em crossmodal triplet training} systems (b), with the {\bf RECOLA} database for {\bf valence} regression. The best performed $\alpha$ or $\beta$ is indicated in each case.}
    \label{fig:res_weights_recola_triplet}
\end{figure}

{\blue Fig.~\ref{fig:res_weights_recola_joint} (a) and Fig.~\ref{fig:res_weights_recola_triplet} (a) depict the relationship between the obtained CCCs and the weight $\alpha$ (cf.~Eq.~(\ref{eq:fuse1}) and (\ref{eq:fuse2})) on the RECOLA database. It is noted that the model performance is improved when the weight increases to some values for the video-based arousal regression models (green and cyan lines in Fig.~\ref{fig:res_weights_recola_joint} (a)). 
Similar observations can be made as well for the audio-based valence regression models (blue and red lines in Fig.~\ref{fig:res_weights_recola_triplet} (a)).}
Therefore, this behaviour again indicates that learning from other modalities indeed can help the enhancement of traditional monomodal systems. Yet, it is also noted that the audio-based arousal and the video-based valence regression models almost remain without obvious performance improvement. This might suggest that transferring the information from the modality with richer knowledge to the one with sparse knowledge is much easier than the other way around, as audio signals often lead to higher CCCs for arousal regression while video signals for valence regression. 

{\blue Further, Fig.~\ref{fig:res_weights_recola_joint} (b) and Fig.~\ref{fig:res_weights_recola_triplet} (b) illustrate the relationship between the obtained CCCs and the weight $\beta$ (cf.~Eq.~(\ref{eq:triplet})) on the RECOLA database. Obviously, it can be seen that the obtained CCCs remarkably grow with the increase of weight $\beta$ in all cases for arousal (cf.~Fig.~\ref{fig:res_weights_recola_joint} (b)) and valence (cf.~Fig.~\ref{fig:res_weights_recola_triplet} (b)) regression.} Specifically, when $\beta=1.0$, \ie the triplet training contributes equally as the traditional emotion regression training, the systems yield the best CCCs in all cases for arousal regression. Nevertheless, the audio- and video-based valence regression systems deliver the best CCCs only when $\beta=0.4$ and $\beta=0.8/1.2$, respectively. The lower contribution from triplet loss implies that it might be more difficult to distil the valence-salient representations than the arousal-salient representations by means of triplet training.  

\begin{figure}[!t]
    \centering
    \subfigure[joint auxiliary modality loss]{
    \includegraphics[height=1.3in,width=1.65in,trim={0cm 0cm 0cm 0cm}, clip]{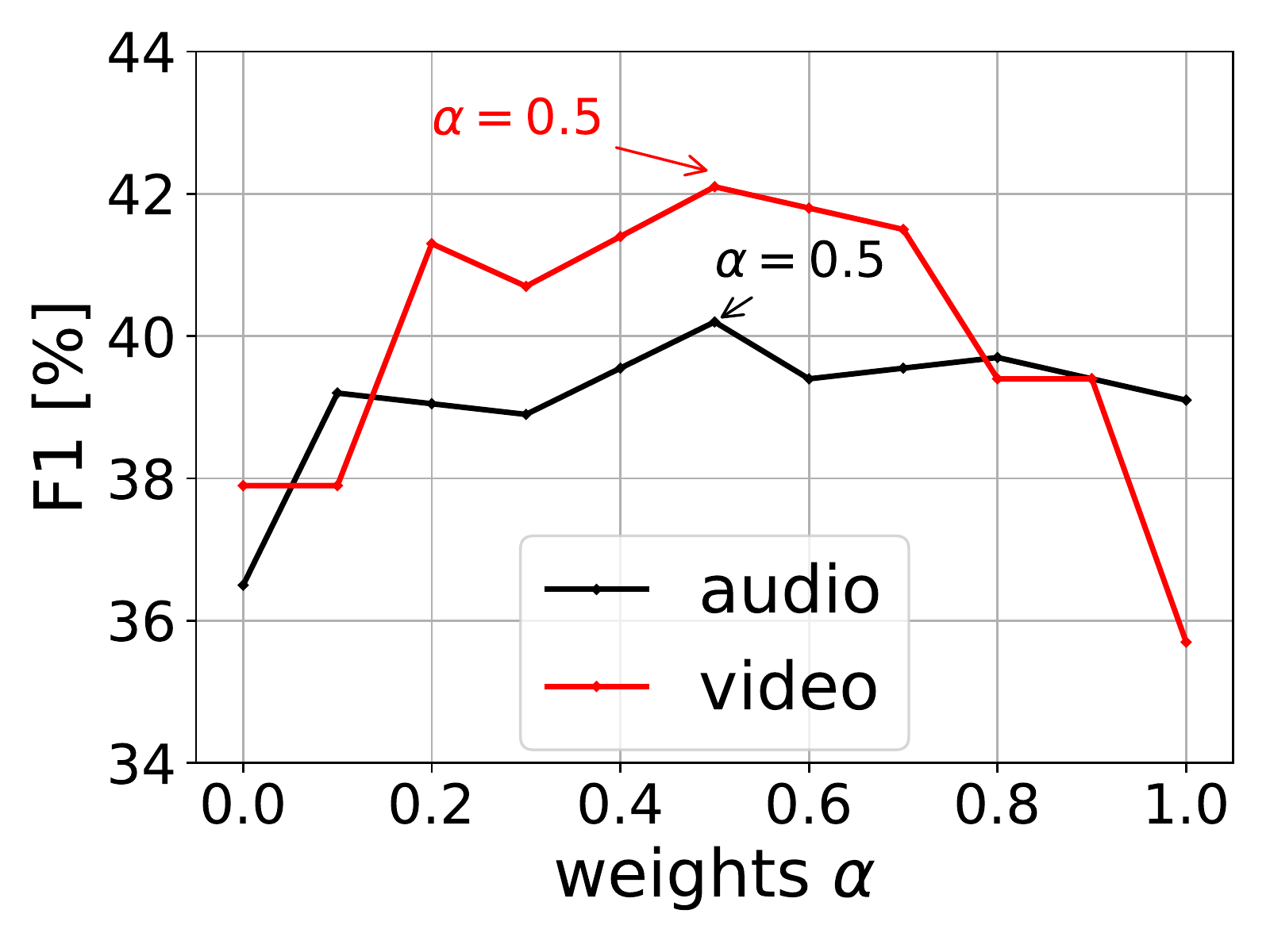}
    }
    \subfigure[crossmodal triplet loss]{
    \includegraphics[height=1.3in,width=1.65in,trim={0cm 0cm 0cm 0cm}, clip]{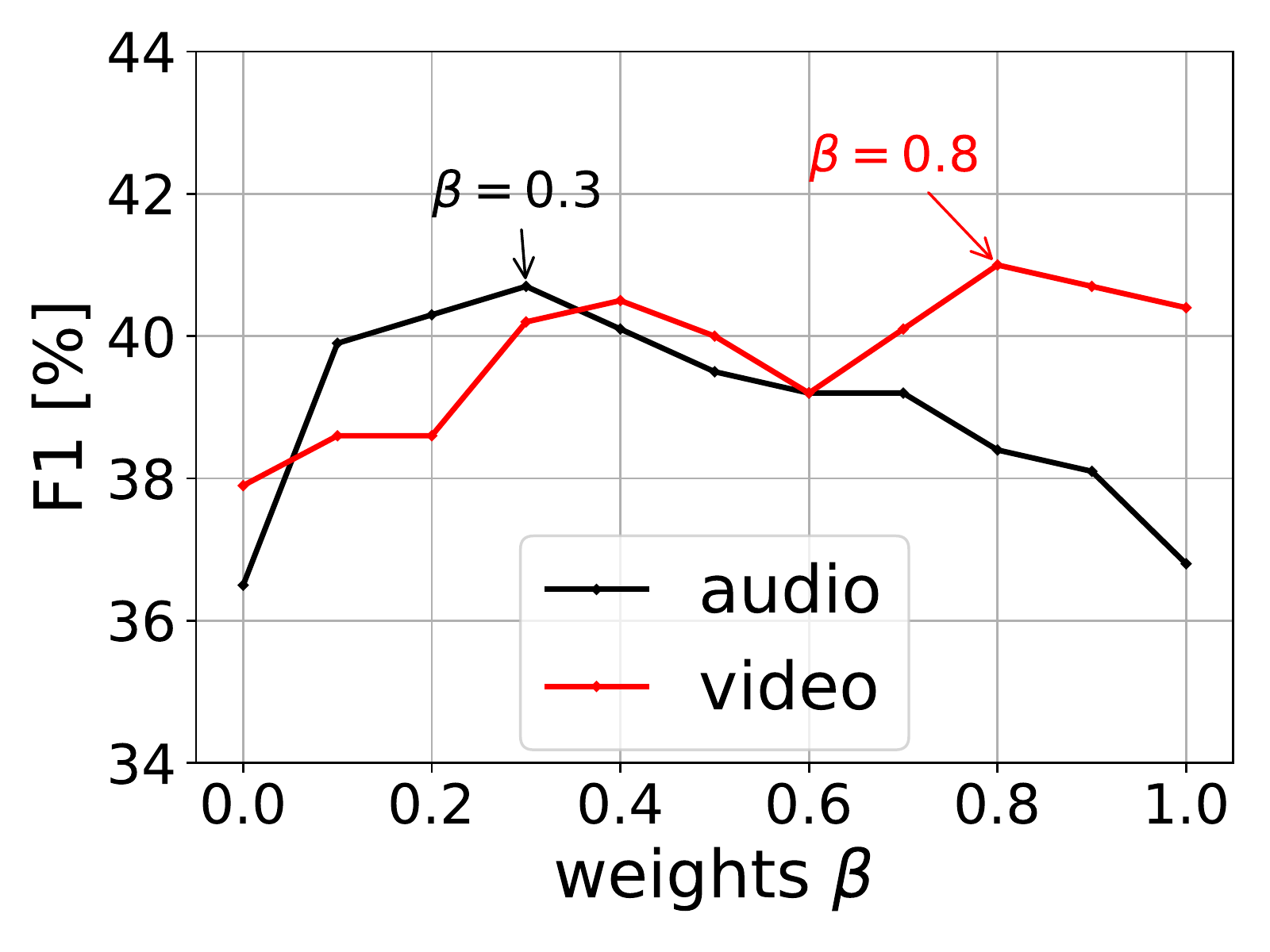}
    }
    \caption{\blue Impact of the joint auxiliary modality loss on the {\em joint audiovisual training} systems (a), and impact of the crossmodal triplet loss on the {\em crossmodal triplet training} systems (b), with the {\bf OMG-Emotion} database. The best performed $\alpha$ or $\beta$ is indicated in each case.}
    \label{fig:res_weights_omg}
\end{figure}

Moreover, we conducted a similar investigation on the OMG-Emotion database for categorical emotion classification. Fig.~\ref{fig:res_weights_omg} explicitly quantifies the contributions of joint audiovisual training (a) and crossmodal triplet training (b) when in a crossmodal learning framework. 
For a joint audiovisual training system, when $\alpha = 0.0$, \ie no contribution from the auxiliary modality, the model is learnt based on only the loss of each modality, separately. When $\alpha$ increases, \ie the contribution of the auxiliary modality during training increases, the performance of monomodal emotion recognition (audio or video) is improved first, until a point where the contribution of the auxiliary modality might actually penalise the learning objective too much and even harm the learning of the main modality, and thus performances start to decrease. Similar observations can be found for crossmodal triplet training systems. 

To this end, proper values of the weight $\alpha$ and $\beta$ need to be identified for the tasks at hand. We can observe from the figure that, the best performance for both audio and video emotion classification is reached when $\alpha = 0.5$ in joint audiovisual training systems; whereas the best performance for audio and video emotion classification is achieved when $\beta=0.3$ and $\beta=0.8$, respectively, in crossmodal triplet training systems.

\section{Conclusion}
\label{sec:conclusion}
Different from previous emotion recognition works which have focused on either the traditional monomodal (\ie modality-specific) systems or multimodal systems, in this article, we proposed an enhanced system through exploring the information across auxiliary modalities. To implement this system with {\blue exemplary} audio and visual modalities, we, on the one hand, utilised a shared emotion recognition network for both audio and video data, so that the complementary information from an auxiliary modality can be implicitly transferred to the target modality. On the other hand, we applied a triplet constraint over acoustic and visual representations to distil emotional embeddings that are invariant to the modalities. The proposed learning frameworks were systematically evaluated on the two benchmark databases RECOLA and OMG-Emotion. Experimental results have demonstrated that the proposed methods significantly improve the prediction performance of a monomodal system, by fusing an additional modality in the training process. 

Albeit the efficiency, the proposed learning framework could be further developed in the future. 
For example, in the triplet training process, the annotation uncertainty information could be utilised as a new distance measure between the learnt representations. In addition, it is also worth to train the model by using  large-scale heterogeneous datasets from a variety of domains. Moreover, in this work, we independently conducted the feature extraction process before performing the crossmodal learning process. It might be helpful if we further combine the two processes together as an end-to-end framework.

\section*{Acknowledgments}
This work was supported by the TransAtlantic Platform ``Digging into Data'' collaboration grant (ACLEW: Analysing Child Language Experiences Around The World), with the support of the UK's Economic \& Social Research Council through the research Grant No.~HJ-253479, and by the EU's Horizon H2020 research and innovation programme under Marie Sk\l{}odowska-Curie grant agreement No.\,766287 (TAPAS).

\ifCLASSOPTIONcaptionsoff
  \newpage
\fi

\bibliographystyle{IEEEtran}
\bibliography{ref}

\begin{thebibliography}{10}
\providecommand{\url}[1]{#1}
\csname url@samestyle\endcsname
\providecommand{\newblock}{\relax}
\providecommand{\bibinfo}[2]{#2}
\providecommand{\BIBentrySTDinterwordspacing}{\spaceskip=0pt\relax}
\providecommand{\BIBentryALTinterwordstretchfactor}{4}
\providecommand{\BIBentryALTinterwordspacing}{\spaceskip=\fontdimen2\font plus
\BIBentryALTinterwordstretchfactor\fontdimen3\font minus
  \fontdimen4\font\relax}
\providecommand{\BIBforeignlanguage}[2]{{%
\expandafter\ifx\csname l@#1\endcsname\relax
\typeout{** WARNING: IEEEtran.bst: No hyphenation pattern has been}%
\typeout{** loaded for the language `#1'. Using the pattern for}%
\typeout{** the default language instead.}%
\else
\language=\csname l@#1\endcsname
\fi
#2}}
\providecommand{\BIBdecl}{\relax}
\BIBdecl

\bibitem{Cowie01-Emotion}
R.~Cowie, E.~Douglas-Cowie, N.~Tsapatsoulis, G.~Votsis, S.~Kollias, W.~Fellenz,
  and J.~G. Taylor, ``Emotion recognition in human-computer interaction,''
  \emph{IEEE Signal Processing Magazine}, vol.~18, no.~1, pp. 32--80, Jan.
  2001.

\bibitem{Zeng09-Survey}
Z.~Zeng, M.~Pantic, G.~I. Roisman, and T.~S. Huang, ``A survey of affect
  recognition methods: {Audio}, visual, and spontaneous expressions,''
  \emph{IEEE Transactions on Pattern Analysis and Machine Intelligence},
  vol.~31, no.~1, pp. 39--58, Jan. 2009.

\bibitem{Zhang14-Distributing}
Z.~Zhang, E.~Coutinho, J.~Deng, and B.~Schuller, ``Distributing recognition in
  computational paralinguistics,'' \emph{IEEE Transactions on Affective
  Computing}, vol.~5, no.~4, pp. 406--417, Oct. 2014.

\bibitem{Cambria16-Affective}
E.~Cambria, ``Affective computing and sentiment analysis,'' \emph{IEEE
  Intelligent Systems}, vol.~31, no.~2, pp. 102--107, Mar. 2016.

\bibitem{El11-Survey}
M.~El~Ayadi, M.~S. Kamel, and F.~Karray, ``Survey on speech emotion
  recognition: {Features}, classification schemes, and databases,''
  \emph{Pattern Recognition}, vol.~44, no.~3, pp. 572--587, Mar. 2011.

\bibitem{Happy15-Automatic}
S.~L. Happy and A.~Routray, ``Automatic facial expression recognition using
  features of salient facial patches,'' \emph{IEEE Transactions on Affective
  Computing}, vol.~6, no.~1, pp. 1--12, Jan. 2015.

\bibitem{Clavel16-Sentiment}
C.~Clavel and Z.~Callejas, ``Sentiment analysis: {F}rom opinion mining to
  human-agent interaction,'' \emph{IEEE Transactions on Affective Computing},
  vol.~7, no.~1, pp. 74--93, Jan. 2016.

\bibitem{Alarcao18-Emotions}
S.~M. Alarcao and M.~J. Fonseca, ``Emotions recognition using {EEG} signals:
  {A} survey,'' \emph{IEEE Transactions on Affective Computing}, June 2018, 20
  pages.

\bibitem{Zhang18-Cross}
B.~Zhang, E.~M. Provost, and G.~Essl, ``Cross-corpus acoustic emotion
  recognition with multi-task learning: Seeking common ground while preserving
  differences,'' \emph{IEEE Transactions on Affective Computing}, Mar. 2017, 14
  pages.

\bibitem{Han18-Adversarial}
J.~Han, Z.~Zhang, N.~Cummins, and B.~Schuller, ``Adversarial training in
  affective computing and sentiment analysis: {Recent} advances and
  perspectives,'' \emph{IEEE Computational Intelligence Magazine}, 2018, 13
  pages.

\bibitem{Mohammed18-DAF}
M.~Abdelwahab and C.~Busso, ``Domain adversarial for acoustic emotion
  recognition,'' \emph{IEEE/ACM Transactions on Audio, Speech, and Language
  Processing}, vol.~26, no.~12, pp. 2423--2435, Dec. 2018.

\bibitem{Nicolaou11-Continuous}
M.~Nicolaou, H.~Gunes, and M.~Pantic, ``Continuous prediction of spontaneous
  affect from multiple cues and modalities in valence-arousal space,''
  \emph{IEEE Transactions on Affective Computing}, vol.~2, no.~2, pp. 92--105,
  Apr. 2011.

\bibitem{Poria16-Fusing}
S.~Poria, E.~Cambria, N.~Howard, G.-B. Huang, and A.~Hussain, ``Fusing audio,
  visual and textual clues for sentiment analysis from multimodal content,''
  \emph{Neurocomputing}, vol. 174, pp. 50--59, Jan. 2016.

\bibitem{Han17-Strength}
J.~Han, Z.~Zhang, N.~Cummins, F.~Ringeval, and B.~Schuller, ``Strength
  modelling for real-world automatic continuous affect recognition from
  audiovisual signals,'' \emph{Image and Vision Computing}, vol.~65, pp.
  76--86, Sep. 2017.

\bibitem{Zhang18-Dynamic}
Z.~Zhang, J.~Han, and B.~Schuller, ``Dynamic difficulty awareness training for
  continuous emotion prediction,'' \emph{IEEE Transactions on Multimedia},
  vol.~PP, Sep. 2018, 13 pages.

\bibitem{Alghowinem18-Multimodal}
S.~Alghowinem, R.~Goecke, M.~Wagner, J.~Epps, M.~Hyett, G.~Parker, and
  M.~Breakspear, ``Multimodal depression detection: Fusion analysis of
  paralinguistic, head pose and eye gaze behaviors,'' \emph{IEEE Transactions
  on Affective Computing}, vol.~9, no.~4, pp. 478--490, Oct. 2018.

\bibitem{Chechik10-Large}
G.~Chechik, V.~Sharma, U.~Shalit, and S.~Bengio, ``Large scale online learning
  of image similarity through ranking,'' \emph{Journal of Machine Learning
  Research}, vol.~11, no. Mar., pp. 1109--1135, 2010.

\bibitem{Schroff15-Facenet}
F.~Schroff, D.~Kalenichenko, and J.~Philbin, ``{FaceNet: A} unified embedding
  for face recognition and clustering,'' in \emph{Proc.\ IEEE conference on
  Computer Vision and Pattern Recognition (CVPR)}, Boston, MA, 2015, pp.
  815--823.

\bibitem{Eyben12-MAC}
F.~Eyben, M.~W\"{o}llmer, and B.~Schuller, ``A multitask approach to continuous
  five-dimensional affect sensing in natural speech,'' \emph{ACM Transactions
  on Interactive Intelligent Systems}, vol.~2, no.~1, pp. 1--29, Mar. 2012.

\bibitem{Xia17-AMT}
R.~Xia and Y.~Liu, ``A multi-task learning framework for emotion recognition
  using {2D} continuous space,'' \emph{IEEE Transactions on Affective
  Computing}, vol.~8, no.~1, pp. 3--14, Jan. 2017.

\bibitem{Han17-From}
J.~Han, Z.~Zhang, M.~Schmitt, and B.~Schuller, ``From hard to soft: {T}owards
  more human-like emotion recognition by modelling the perception
  uncertainty,'' in \emph{Proc.\ ACM International Conference on Multimedia
  (MM)}, Mountain View, CA, 2017, pp. 890--897.

\bibitem{Zhao18-MMM}
J.~Zhao, R.~Li, S.~Chen, and Q.~Jin, ``Multi-modal multi-cultural dimensional
  continues emotion recognition in dyadic interactions,'' in \emph{Proc.\ 8th
  International Workshop on Audio/Visual Emotion Challenge (AVEC)}, Seoul,
  South Korea, 2018, pp. 65--72.

\bibitem{Taylor18-Personalized}
S.~A. Taylor, N.~Jaques, E.~Nosakhare, A.~Sano, and R.~Picard, ``Personalized
  multitask learning for predicting tomorrow's mood, stress, and health,''
  \emph{IEEE Transactions on Affective Computing}, Dec. 2018, 14 pages.

\bibitem{Han18-Emotion}
J.~Han, Z.~Zhang, G.~Keren, and B.~Schuller, ``Emotion recognition in speech
  with latent discriminative representations learning,'' \emph{Acta Acustica
  united with Acustica}, vol. 104, no.~5, pp. 737--740, Sep. 2018.

\bibitem{Huang18-Speech}
J.~Huang, Y.~Li, J.~Tao, and Z.~Lian, ``Speech emotion recognition from
  variable-length inputs with triplet loss function,'' in \emph{Proc.\ Annual
  Conference of the International Speech Communication Association
  (INTERSPEECH)}, Hyderabad, India, 2018, pp. 3673--3677.

\bibitem{Yang15-CFL}
X.~Yang, T.~Zhang, and C.~Xu, ``Cross-domain feature learning in multimedia,''
  \emph{IEEE Transactions on Multimedia}, vol.~17, no.~1, pp. 64--78, Jan.
  2015.

\bibitem{Kang15-LCF}
C.~Kang, S.~Xiang, S.~Liao, C.~Xu, and C.~Pan, ``Learning consistent feature
  representation for cross-modal multimedia retrieval,'' \emph{IEEE
  Transactions on Multimedia}, vol.~17, no.~3, pp. 370--381, Mar. 2015.

\bibitem{Aytar16-SLS}
Y.~Aytar, C.~Vondrick, and A.~Torralba, ``{SoundNet}: Learning sound
  representations from unlabeled video,'' in \emph{Proc.\ Advances in Neural
  Information Processing Systems (NIPS)}, Barcelona, Spain, 2016, pp. 892--900.

\bibitem{Wang17-ACM}
B.~Wang, Y.~Yang, X.~Xu, A.~Hanjalic, and H.~T. Shen, ``Adversarial cross-modal
  retrieval,'' in \emph{Proc.\ ACM International Converence on Multimedia
  (MM)}, Mountain View, CA, 2017, pp. 154--162.

\bibitem{Albanie18-ERS}
S.~Albanie, A.~Nagrani, A.~Vedaldi, and A.~Zisserman, ``Emotion recognition in
  speech using cross-modal transfer in the wild,'' in \emph{Proc.\ ACM
  International Conference on Multimedia (MM)}, Seoul, Korea, 2018, pp.
  292--301.

\bibitem{Metallinou12-Context}
A.~Metallinou, M.~W\"ollmer, A.~Katsamanis, F.~Eyben, B.~Schuller, and
  S.~Narayanan, ``Context-sensitive learning for enhanced audiovisual emotion
  classification,'' \emph{IEEE Transactions on Affective Computing}, vol.~3,
  no.~2, pp. 184--198, Jan. 2012.

\bibitem{Busso04-AOE}
C.~Busso, Z.~Deng, S.~Yildirim, M.~Bulut, C.~M. Lee, A.~Kazemzadeh, S.~Lee,
  U.~Neumann, and S.~Narayanan, ``Analysis of emotion recognition using facial
  expressions, speech and multimodal information,'' in \emph{Proc.\
  International Conference on Multimodal Interfaces (ICMI)}, State College, PA,
  2004, pp. 205--211.

\bibitem{Wimmer08-LFO}
M.~Wimmer, B.~Schuller, D.~Arsic, B.~Radig, and G.~Rigoll, ``Low-level fusion
  of audio and video feature for multi-modal emotion recognition,'' in
  \emph{Proc.\ International Conference on Computer Vision Theory and
  Applications (VISAPP)}, Funchal, Portugal, 2008, pp. 145--151.

\bibitem{Metallinou10-DLC}
A.~Metallinou, S.~Lee, and S.~Narayanan, ``Decision level combination of
  multiple modalities for recognition and analysis of emotional expression,''
  in \emph{Proc.\ IEEE International Conference on Acoustics, Speech and Signal
  Processing (ICASSP)}, Dallas, TX, 2010, pp. 2462--2465.

\bibitem{Zeng05-AAR}
Z.~Zeng, J.~Tu, B.~Pianfetti, M.~Liu, T.~Zhang, Z.~Zhang, T.~S. Huang, and
  S.~Levinson, ``Audio-visual affect recognition through multi-stream fused
  {HMM} for {HCI},'' in \emph{Proc.\ IEEE Conference on Computer Vision and
  Pattern Recognition (CVPR)}, San Diego, CA, 2005, pp. 967--972.

\bibitem{Han17-Prediction}
J.~Han, Z.~Zhang, F.~Ringeval, and B.~Schuller, ``Prediction-based learning for
  continuous emotion recognition in speech,'' in \emph{Proc.\ IEEE
  International Conference on Acoustics, Speech and Signal Processing
  (ICASSP)}, New Orleans, LA, 2017, pp. 5005--5009.

\bibitem{Toprak18-Evaluating}
S.~Toprak, N.~Navarro-Guerrero, and S.~Wermter, ``Evaluating integration
  strategies for visuo-haptic object recognition,'' \emph{Cognitive
  computation}, vol.~10, no.~3, pp. 408--425, June 2018.

\bibitem{hoffer15-DML}
E.~Hoffer and N.~Ailon, ``Deep metric learning using triplet network,'' in
  \emph{Prof.\ International Workshop on Similarity-Based Pattern Recognition
  (SIMBAD)}, Copenhagen, Denmark, 2015, pp. 84--92.

\bibitem{huang2013learning}
P.~Huang, X.~He, J.~Gao, L.~Deng, A.~Acero, and L.~Heck, ``Learning deep
  structured semantic models for web search using clickthrough data,'' in
  \emph{Proc.\ International Conference on Information and Knowledge Management
  (CIKM)}, San Francisco, CA, 2013, pp. 2333--2338.

\bibitem{Ding18-TEC}
C.~Ding and D.~Tao, ``Trunk-branch ensemble convolutional neural networks for
  video-based face recognition,'' \emph{IEEE Transactions on Pattern Analysis
  and Machine Intelligence}, vol.~40, no.~4, pp. 1002--1014, Apr. 2018.

\bibitem{Valstar16-AVEC}
M.~Valstar, J.~Gratch, B.~Schuller, F.~Ringeval, D.~Lalanne, M.~Torres~Torres,
  S.~Scherer, G.~Stratou, R.~Cowie, and M.~Pantic, ``{AVEC} 2016: Depression,
  mood, and emotion recognition workshop and challenge,'' in \emph{Proc.\ 6th
  International Workshop on Audio/Visual Emotion Challenge (AVEC)}, Amsterdam,
  The Netherlands, 2016, pp. 3--10.

\bibitem{Ringeval18-AVEC}
F.~Ringeval, B.~Schuller, M.~Valstar, R.~Cowie, H.~Kaya, M.~Schmitt,
  S.~Amiriparian, N.~Cummins, D.~Lalanne, A.~Michaud, E.~Cift\c{c}i,
  H.~G\"{u}le\c{c}, A.~Salah, and M.~Pantic, ``{AVEC} 2018 workshop and
  challenge: Bipolar disorder and cross-cultural affect recognition,'' in
  \emph{Proc.\ 8th International Workshop on Audio/Visual Emotion Challenge
  (AVEC)}, Seoul, South Korea, 2018, pp. 3--13.

\bibitem{Ringeval13-Introducing}
F.~Ringeval, A.~Sonderegger, J.~S. Sauer, and D.~Lalanne, ``Introducing the
  \textsc{RECOLA} multimodal corpus of remote collaborative and affective
  interactions,'' in \emph{Proc.\ 10th {IEEE} International Conference and
  Workshops on Automatic Face and Gesture Recognition (FG)}, Shanghai, China,
  2013, pp. 1--8.

\bibitem{Ringeval15-AV+EC}
F.~Ringeval, B.~Schuller, M.~Valstar, S.~Jaiswal, E.~Marchi, D.~Lalanne,
  R.~Cowie, and M.~Pantic, ``{AV+EC} 2015: The first affect recognition
  challenge bridging across audio, video, and physiological data,'' in
  \emph{Proc.\ 5th International Workshop on Audio/Visual Emotion Challenge
  (AVEC)}, Brisbane, Australia, 2015, pp. 3--8.

\bibitem{Eyben16-Geneva}
F.~Eyben, K.~Scherer, B.~Schuller, J.~Sundberg, E.~Andr\'e, C.~Busso,
  L.~Devillers, J.~Epps, P.~Laukka, S.~Narayanan, and K.~Truong, ``{The Geneva
  Minimalistic Acoustic Parameter Set} ({GeMAPS}) for voice research and
  affective computing,'' \emph{IEEE Transactions on Affective Computing},
  vol.~7, no.~2, pp. 190--202, Apr. 2016.

\bibitem{Eyben10-OTM}
F.~Eyben, M.~W\"ollmer, and B.~Schuller, ``{openSMILE} -- {T}he {M}unich
  versatile and fast open-source audio feature extractor,'' in \emph{Proc.\ ACM
  International Conference on Multimedia (MM)}, Florence, Italy, 2010, pp.
  1459--1462.

\bibitem{Barros18-OMG}
P.~Barros, N.~Churamani, E.~Lakomkin, H.~Siqueira, A.~Sutherland, and
  S.~Wermter, ``The {OMG}-emotion behavior dataset,'' in \emph{Proc.\
  International Joint Conference on Neural Networks (IJCNN)}, Rio, Brazil,
  2018, pp. 1408--1412.

\bibitem{Zhang16-JFD}
K.~Zhang, Z.~Zhang, Z.~Li, and Y.~Qiao, ``Joint face detection and alignment
  using multitask cascaded convolutional networks,'' \emph{IEEE Signal
  Processing Letters}, vol.~23, no.~10, pp. 1499--1503, Oct. 2016.

\bibitem{Parkhi15-DFR}
O.~M. Parkhi, A.~Vedaldi, and A.~Zisserman, ``Deep face recognition,'' in
  \emph{Proc.\ British Machine Vision Conference}, Swansea, UK, 2015, pp.
  1--12.

\bibitem{Cho14-properties}
K.~Cho, B.~Van~Merri{\"e}nboer, D.~Bahdanau, and Y.~Bengio, ``On the properties
  of neural machine translation: {Encoder-decoder} approaches,'' in
  \emph{Proc.\ Workshop on Syntax, Semantics and Structure in Statistical
  Translation (SSST)}, Doha, Qatar, 2014, pp. 103--111.

\bibitem{Jozefowicz15-empirical}
R.~Jozefowicz, W.~Zaremba, and I.~Sutskever, ``An empirical exploration of
  recurrent network architectures,'' in \emph{Proc.\ International Conference
  on Machine Learning (ICML)}, Lille, France, 2015, pp. 2342--2350.

\bibitem{Mariooryad15-Correcting}
S.~Mariooryad and C.~Busso, ``Correcting time-continuous emotional labels by
  modeling the reaction lag of evaluators,'' \emph{IEEE Transactions on
  Affective Computing}, vol.~6, no.~2, pp. 97--108, Apr. 2015.

\bibitem{Cohen13-Applied}
J.~Cohen, P.~Cohen, S.~G. West, and L.~S. Aiken, \emph{Applied multiple
  regression/correlation analysis for the behavioral sciences}.\hskip 1em plus
  0.5em minus 0.4em\relax Abingdon, UK: Routledge, 2013.

\bibitem{Tzirakis17-End}
P.~Tzirakis, G.~Trigeorgis, M.~A. Nicolaou, B.~Schuller, and S.~Zafeiriou,
  ``End-to-end multimodal emotion recognition using deep neural networks,''
  \emph{IEEE Journal of Selected Topics in Signal Processing, Special Issue on
  End-to-End Speech and Language Processing}, vol.~11, no.~8, pp. 1301--1309,
  Dec. 2017.

\bibitem{Han17-Reconstruction}
J.~Han, Z.~Zhang, F.~Ringeval, and B.~Schuller, ``Reconstruction-error-based
  learning for continuous emotion recognition in speech,'' in \emph{Proc.\ IEEE
  International Conference on Acoustics, Speech and Signal Processing
  (ICASSP)}, New Orleans, LA, 2017, pp. 2367--2371.

\bibitem{Lotfian18-Curriculum}
R.~Lotfian and C.~Busso, ``Curriculum learning for speech emotion recognition
  from crowdsourced labels,'' \emph{arXiv preprint arXiv:1805.10339}, May 2018.

\end{thebibliography}

\vspace{2cm}
\begin{IEEEbiography}[{\includegraphics[width=1in,height=1.25in,clip,keepaspectratio]{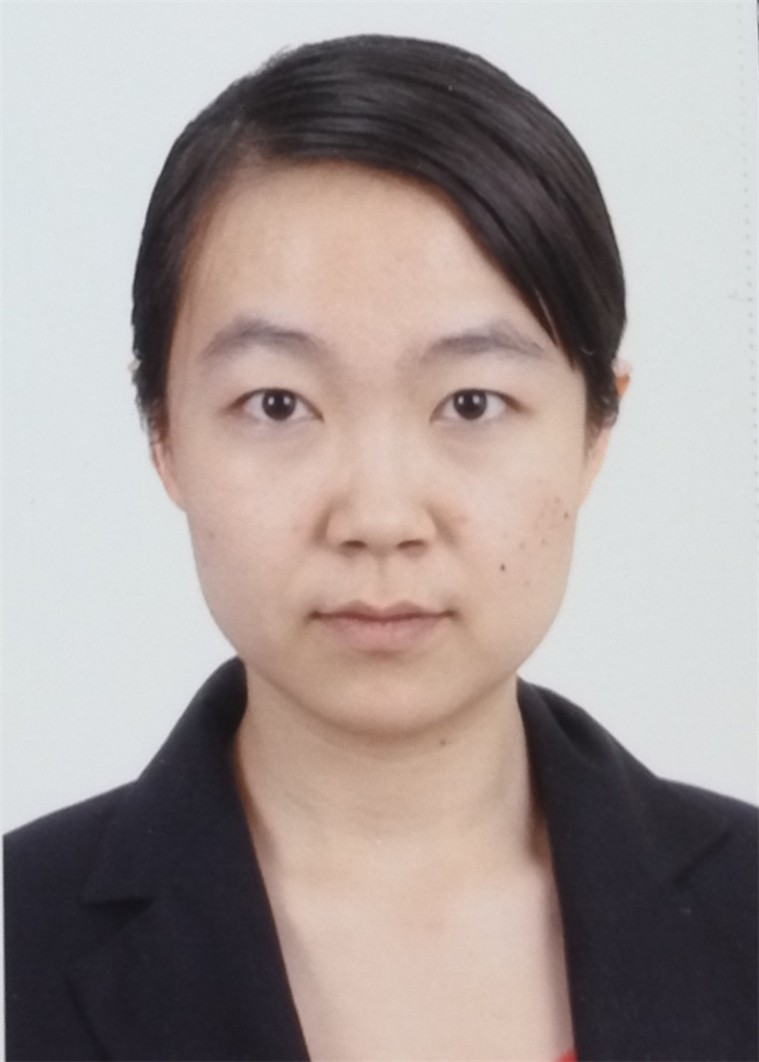}}]
{\bf Jing Han} (S'16) received her bachelor degree (2011) in electronic and information engineering from Harbin Engineering University (HEU), China, and her master degree (2014) from Nanyang Technological University, Singapore. She is now working as a doctoral student with the ZD.B Chair of Embedded Intelligence for Health Care and Wellbeing at the University of Augsburg, Germany, involved in two EU's Horizon 2020 projects SEWA and RADAR CNS. She reviews regularly for IEEE Transactions on Cybernetics, IEEE Access, and IEEE Signal Processing Letters. Her research interests are related to deep learning for multimodal affective computing and health care. Besides, she co-chaired the 7th Audio/Visual Emotion Challenge (AVEC) and workshop in 2017, and served as a program committee member of the 8th AVEC challenge and workshop in 2018. Moreover, she was awarded student travel grants from IEEE SPS and ISCA to attend ICASSP and INTERSPEECH in 2018.
\end{IEEEbiography}

\vspace{-1cm}
\begin{IEEEbiography}[{\includegraphics[width=1in,height=1.25in,clip,keepaspectratio]{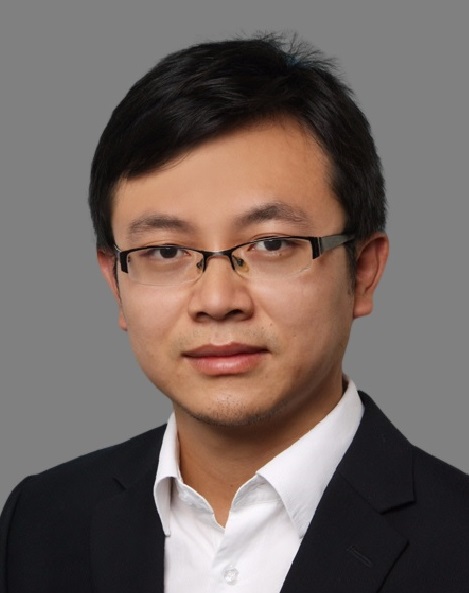}}]
{\bf Zixing Zhang} (M'15) received his master degree in physical electronics from the Beijing University of Posts and Telecommunications (BUPT), China, in 2010, and his PhD degree in computer engineering from Technical University of Munich (TUM), Germany, in 2015. Currently, he is a research associate with the Department of Computing at the Imperial College London (ICL), UK, since 2017. Before that, he was a postdoctoral researcher at the University of Passau, Germany, from 2015 to 2017. He has (co-)authored more than 80 publications in peer-reviewed books, journals, and conference proceedings to date. 
His research mainly focuses on deep learning technologies for the speaker-centred state and health computing. He has organised special sessions, such as at the IEEE 7th Affective Computing and Intelligent Interaction (ACII) conference in 2017 and at the 43nd IEEE International Conference on Acoustics, Speech, and Signal Processing (ICASSP) in 2018. Moreover, he serves as a reviewer for numerous leading-in-their fields journals and conferences, a programme committee member and an area chair for many international conferences, and a leading guest editor of the IEEE Transactions on Emerging Topics in Computational Intelligence. 
\end{IEEEbiography}

\vspace{-1cm}
\begin{IEEEbiography}[{\includegraphics[width=1in,height=1.25in,clip,keepaspectratio]{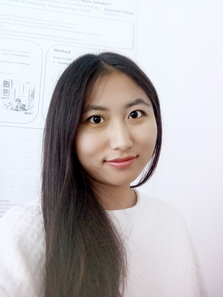}}] 
{Zhao Ren} received her master degree in Computer Science and Technology from the Northwestern Polytechnical University (NWPU) in the P.R.~China, 2017. Currently, she is a EU-researcher and working on her Ph.\,D.~degree at the ZD.B Chair of Embedded Intelligence for Health Care and Wellbeing, University of Augsburg, Germany, where she is involved with the H2020-MSCA-ITN-ETN project TAPAS, for health-related  speech analysis. Her research interests mainly lie in transfer learning, unsupervised learning, and deep learning for the application in health care and wellbeing.
\end{IEEEbiography}

\vspace{-1cm}
\begin{IEEEbiography}[{\includegraphics[width=1in,height=1.25in,clip,keepaspectratio]{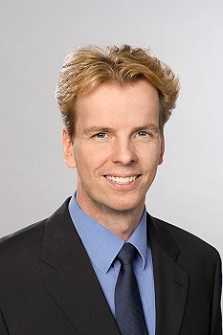}}] 
{\bf Bj\"orn~Schuller} (M'05-SM'15-F'18) received his diploma in 1999, his doctoral degree for his study on Automatic Speech and Emotion Recognition in 2006, and his habilitation and Adjunct Teaching Professorship in the subject area of Signal Processing and Machine Intelligence in 2012, all in electrical engineering and information technology from TUM in Munich/Germany. He is Professor of Artificial Intelligence in the Department of Computing at the Imperial College London/UK, where he heads GLAM -- the Group on Language, Audio \& Music, Full Professor and head of the ZD.B Chair of Embedded Intelligence for Health Care and Wellbeing at the University of Augsburg/Germany, and CEO of audEERING.  
He was previously full professor and head of the Chair of Complex and Intelligent Systems at the University of Passau/Germany. 
Professor Schuller is President-emeritus of the Association for the Advancement of Affective Computing (AAAC), elected member of the IEEE Speech and Language Processing Technical Committee, Fellow of the IEEE, and Senior Member of the ACM. He (co-)authored 5 books and more than 800 publications in peer-reviewed books, journals, and conference proceedings leading to more than overall 23\,000 citations (h-index = 73). Schuller is general chair of ACII 2019, co-Program Chair of Interspeech 2019 and ICMI 2019, repeated Area Chair of ICASSP, and former Editor in Chief of the IEEE Transactions on Affective Computing next to a multitude of further Associate and Guest Editor roles and functions in Technical and Organisational Committees.\end{IEEEbiography}

\end{document}